\pdfoutput=1
\documentclass[11pt]{article}

\usepackage[final]{acl}
\usepackage{times}
\usepackage{latexsym}

\usepackage[T1]{fontenc}
\usepackage[utf8]{inputenc}

\usepackage{microtype}

\usepackage{inconsolata}

\usepackage{hyperref}
\usepackage{url}
\usepackage{array}
\usepackage{booktabs}
\usepackage{amsmath}
\usepackage{amssymb}
\usepackage{graphicx}
\usepackage{multirow}
\usepackage{float}
\usepackage{algorithm}
\usepackage{algpseudocode}
\usepackage{subcaption}
\usepackage{longtable}
\usepackage{colortbl}
\usepackage{cleveref}

\usepackage{amsthm}

\newtheorem{definition}{Definition}[section]
\newtheorem{proposition}{Proposition}[section]
\newtheorem{property}{Property}[section]

\title{Permutative Preference Alignment\\ from Listwise Ranking of Human Judgments}

\author{Yang Zhao$^{1\dagger}$ \quad Yixin Wang$^2$ \quad Mingzhang Yin$^{3\dagger}$%\thanks{Corresponding author.} 
\\$^1$University of Texas at Austin \quad $^2$University of Michigan, Ann Arbor \quad $^3$University of Florida\\
\texttt{yangzhao25@utexas.edu \quad yixinw@umich.edu}
\\
\texttt{mingzhang.yin@warrington.ufl.edu}}

\begin{document}
\maketitle

\renewcommand{\thefootnote}{\fnsymbol{footnote}}
\footnotetext[2]{\;Corresponding author.}

\begin{abstract}
Aligning Large Language Models (LLMs) with human preferences is crucial in ensuring desirable and controllable model behaviors. Current methods, such as Reinforcement Learning from Human Feedback (RLHF) and Direct Preference Optimization (DPO), rely on the Bradley-Terry (B-T) model to maximize the likelihood of pairwise choices. However, when multiple responses are available, the B-T model fails to guarantee an accurate list ranking of the responses. To address this issue, we propose Permutative Preference Alignment (PPA), a novel offline listwise approach that incorporates the Normalized Discounted Cumulative Gain (NDCG)—a widely-used ranking metric—as an alternative training objective for LLM alignment. We develop an end-to-end alignment algorithm by approximating NDCG with a differentiable surrogate loss. Experiments demonstrate that PPA outperforms existing pairwise and listwise methods on evaluation sets and general benchmarks such as AlpacaEval. Furthermore, we show that NDCG-based approaches improve ranking accuracy more effectively than B-T-based methods and provide a theoretical explanation for this improvement.

%by pairwise comparisons. However, when multiple responses are available, these approaches fall short of leveraging the extensive information in the ranking given by the reward models or human feedback. 

%In this work, we propose a novel offline listwise approach named Permutative Preference Alignment (PPA), which employs the Normalized Discounted Cumulative Gain (NDCG), a widely-used ranking metric, to better utilize relative proximity within multiple responses. 
%This approach builds a connection between ranking models in information retrieval and the alignment problem. In aligning multi-response datasets assigned with rewards, PPA outperforms existing pairwise and listwise approaches on evaluation sets and general benchmarks like AlpacaEval. 
\end{abstract}

\section{Introduction}
Large Language Models (LLMs) trained on massive datasets have demonstrated impressive capabilities in natural language processing \citep{gpt4,llama3}. Aligning these models with human preferences is essential for reliable and controllable model behaviors. Pairwise methods, such as Reinforcement Learning from Human Feedback (RLHF) and Direct Preference Optimization (DPO) \citep{dpo}, employ the Bradley-Terry (B-T) model \citep{bradley1952rank}  to maximize the likelihood of pairwise preferences, demonstrating strong performances  \citep{rlhf1,rlhf3}. %The success hinges on the human preferences elicited from pairwise comparisons. 
Various pairwise-based offline preference optimization methods have been developed, such as RRHF \citep{rrhf}, SLiC \citep{slic}, RPO \citep{rpo}, SimPO \citep{simpo}, and LiPO-$\lambda$ \citep{lipo}, which depend on the human preferences elicited from pairwise comparisons.
%primarily modify DPO's reward function and B-T  paradigm. 
These contrastive methods essentially classify preferred and non-preferred responses as positive and negative samples, %aiming to differentiate them as much as possible, 
naturally suited for the binary responses in the data sets like Reddit TL;DR and AnthropicHH \citep{tldr, AnthropicHH}.~\looseness=-1

%However, the RLHF procedure is resource-intensive and sensitive to hyperparameters due to its online multi-stage nature. Direct Preference Optimization (DPO) \citep{dpo}  integrates the multi-stage process into a single offline training objective by eliminating the separate reward model. %which has been widely adopted in practice because of its simplicity and stability.

However, multi-response data are often available, where a single prompt corresponds to several responses with assigned rewards \citep{rlhf3,rrhf, raft,openassistant}. For such data, DPO cannot ensure the correct ranking of individual pairs, as it infers relative quality rankings of responses by maximizing the pairwise choice probability, potentially leading to an inaccurate overall list ranking. Among LTR metrics, NDCG emerges as the ideal training objective because it handles graded relevance and incorporates position-based discounting. Unlike MAP, MRR, Precision, and Recall which suffer from binary relevance limitations and non-differentiability. NDCG \citep{ndcg_recommend} can be effectively approximated with differentiable surrogates. This property makes it uniquely suited for gradient-based alignment optimization.

%Existing pairwise contrastive approaches optimize models by comparing all possible pairs, but they overlook relative proximities of responses. Alternatively, listwise methods present a more comprehensive view of the entire list of responses. Existing listwise methods like DPO-PL, PRO, LIRE \citep{dpo, PRO, LIRE} mainly integrate the Plackett-Luce (PL) model \citep{PL1} to represent the likelihood of list permutations, which is relatively simplistic.

In this work, we propose Permutative Preference Alignment (PPA), a new listwise alignment approach to align human preferences by maximizing NDCG. 
Models can be viewed as score functions that assign reward scores to responses. The alignment is learning to rank these responses to match the permutation derived from ground truth labels. We employ the smooth surrogate loss NeuralNDCG \citep{neuralndcg} to approximate NDCG to overcome its non-differentiable nature. 

\begin{figure*}[ht]
    \centering
    \includegraphics[width=\textwidth]{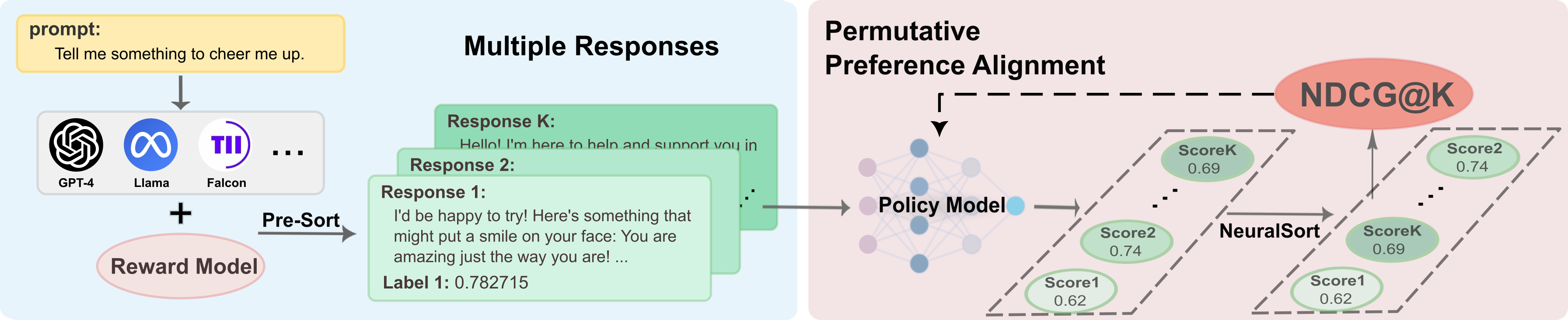}
    \caption{An illustration of Permutative Preference Alignment (PPA) workflow. Each response is assigned a ground truth label by the reward model and pre-sorted in descending order. Reward scores are then derived from the policy and re-sorted to a new permutation. PPA calculates NDCG@K from the difference between two permutations and then optimizes the policy model.}
    \label{fig:ill}
\end{figure*}

In real-world generation tasks, we find that the proposed PPA achieves higher ranking accuracy than B-T model-based methods, a crucial metric in evaluating policy performance. Existing literature shows that B-T-based RLHF and DPO struggle to improve ranking accuracy because they maximize the reward margin between preferred and non-preferred responses  \citep{rank_accuracy}. In contrast, maximizing NDCG only requires the preferred responses to have higher reward scores than the non-preferred ones. Based on this distinction, we provide a theoretical explanation for the improvement of PPA in ranking accuracy. 

%In real-world generation tasks, ranking accuracy is crucial for evaluating policy performance. Studies show that B-T-based RLHF and DPO struggle to improve ranking accuracy, as they maximize the reward margin between preferred and non-preferred responses  \citep{rank_accuracy}. In contrast, maximizing NDCG only requires ensuring preferred responses have higher reward scores than non-preferred ones.
%We show that the proposed PPA achieves higher ranking accuracy than B-T model-based methods and provide a potential theoretical explanation for this observation. 

In empirical studies, we comprehensively evaluate model performance with various pairwise and listwise baselines. In addition, we construct a multiple response dataset assigned with rewards based on UltraFeedback \citep{cui2023ultrafeedback} and SimPO. 
The proposed PPA consistently achieves the best performance on both evaluation datasets and general benchmarks like AlpacaEval \citep{alpaca_eval}.
%This approach aligns LLMs' likelihood closely to human preferences across multi-response datasets, improving the quality of the generative outputs.

Our contributions are summarized as follows:
\begin{itemize}
    \item  We identify potential limitations of DPO in list ranking and introduce NDCG as a training objective to improve ranking performance.
    \item  We propose PPA as a new listwise alignment method that leverages multiple responses, which demonstrates superior performance over existing pairwise and listwise approaches across various model scales.  
    \item We illustrate that NDCG-based method is more effective than Bradley-Terry-based methods in improving ranking accuracy and propose a theoretical explanation. 
\end{itemize}

\section{Related Work}
Recent approaches for aligning language models with human preferences typically fall into three categories. \textbf{Pairwise preference methods} like DPO \citep{dpo} use the Bradley-Terry model to optimize binary preferences without explicit reward models. But in multiple-response scenarios, they focus on average contrastive probability rather than ensuring all individual pairs align with ground truth labels. \textbf{Multiple response alignment} methods like RRHF, LiPO-$\lambda$, DPO-PL, PRO, and LIRE \citep{rrhf,lipo,dpo,PRO,LIRE} expand candidate responses from various LLMs and optimize the model with Bradley-Terry-based or listwise algorithms. Pairwise methods encounter similar limitations as in binary-response conditions, while listwise methods fail to optimize the established evaluation metrics prevalent in the LTR literature, like NDCG. \textbf{Learning to Rank (LTR)} techniques offer promising directions, particularly listwise approaches \citep{xia2008listwise} that consider entire ranking lists as training instances, better capturing response relationships compared to pointwise and pairwise methods. Despite their potential, current listwise techniques have not achieved state-of-the-art performance in LTR, highlighting opportunities for improvement. The further related work details are provided in \Cref{appendix_related_work}.

\section{Preliminaries}
%DPO aligns LLMs with human preferences in a pairwise contrastive manner. In contrast, this paper adopts the Learning to Rank (LTR) framework, which learns how to permute a list of responses. 

Our approach adopts the Learning to Rank (LTR) framework and the list permutations.

\subsection{Problem Setting}
Following the setup in LiPO \citep{lipo}, we assume access to an offline static dataset $\mathcal{D}=\{x^{(i)},\mathbf{Y}^{(i)},\mathbf{\Psi}^{(i)}\}_{i=1}^N$, where $\mathbf{Y}=(y_1,...,y_K)$ is a list of responses from various generative models of size K given the prompt $x$. Each response is associated with a label from $\mathbf{\Psi}=(\psi_1,...,\psi_K)$, also known as the \emph{ground truth labels} in the LTR literature. 
The label $\psi$ measures the quality of responses, which can be generated from human feedback or a pre-trained reward model. 
We obtain the score $\mathbf{\Psi}$ from a reward model:
\begin{align}
    \psi_k=RM(x,y_k),
\label{psi}
\end{align}
where $\psi_k \in [0,1]$. The label is fixed for a response, representing the degree of human preference. 

For each prompt-response pair, we also compute a \emph{reward score} representing the likelihood of the generating probability of the response:
\begin{align}
s_\theta(x,y)=\beta \log \frac{\pi_{\theta}(y|x)}{\pi_{\text{ref}}(y|x)}. 
\label{scores_eq}
\end{align}
Here, $\pi_{\text{ref}}$ is a reference model which we set as the SFT model. 
$\pi_{\theta}(y|x)$ and $\pi_{\text{ref}}(y|x)$ means the probability of the response $y$ given the prompt $x$ under the policy model and the reference model.
Similar to DPO \citep{dpo}, the partition function is omitted due to the symmetry in the choice model of multiple responses. %$Z(x)$ shared by the responses to the same prompt is omitted in computing the reward score for simplicity. 
Unlike the fixed labels $\psi_k$, the 
reward scores $\mathbf{s}=\{s_\theta(x,y_1),...,s_\theta(x,y_K)\}$ depend on the model $\pi_{\theta}$ and are updated during the model training. 

\subsection{NDCG Metric}
NDCG is widely used for evaluating the ranking model performance \citep{ndcg}, which directly assesses the quality of a permutation from the listwise data. Assume the list of responses $\mathbf{Y}=(y_1,...,y_K)$ have been {pre-ranked in the descending order based on labels} $\mathbf{\Psi}=(\psi_1,...,\psi_K)$ from Eq \ref{psi}, where $\psi_i\geq\psi_j$  if $i\geq j$. The Discounted Cumulative Gain at k-th position ($k\leq K$) is defined as:
\begin{equation}
    \text{DCG@}k = \sum_{j=1}^{k} G(\psi_j) D(\tau(j)),
\label{dcg}
\end{equation}
where $\psi_j$ denotes the ground truth labels of the response $y_j$, and $\tau(j)$ is the descending rank position of \(y_j\) based on the reward scores $\mathbf{s}$ computed by the current model $\pi_\theta$. Typically, the discount function and the gain function are set as \(D(\tau(j)) = \frac{1}{\log_2(\tau(j)+1)}\) and \(G(\psi_j) = 2^{\psi_j} - 1\). An illustration is provided in Appendix \ref{appendix_sort}.

The NDCG at k is defined as
\begin{equation}
\text{NDCG@}k = \frac{1}{\text{maxDCG@}k} \text{DCG@}k,
\label{ndcg}
\end{equation}
where \(\text{maxDCG@}k\) is the maximum value of \(\text{DCG@}k\), computed by ordering the responses $\mathbf{Y}$ by their ground truth labels $\mathbf{\Psi}$. The normalization ensures that NDCG is within the range (0, 1). 

The value $k$ of NDCG@$k$ ($k\leq K$) indicates that we focus on the ranking of the top $k$ elements while ignoring those beyond $k$. For example, when $k=2$, we only need to correctly order the first 2 elements, regardless of the order of the remaining $K-2$ elements in the list. %It means solely making $s_1\geq s_2$ (because $\psi_1\geq \psi_2$ always holds) leads to the maximum NDCG@2 value.

\begin{figure*}[ht]
    \centering
    \begin{subfigure}[b]{0.235\textwidth}
        \centering
        \includegraphics[width=\textwidth]{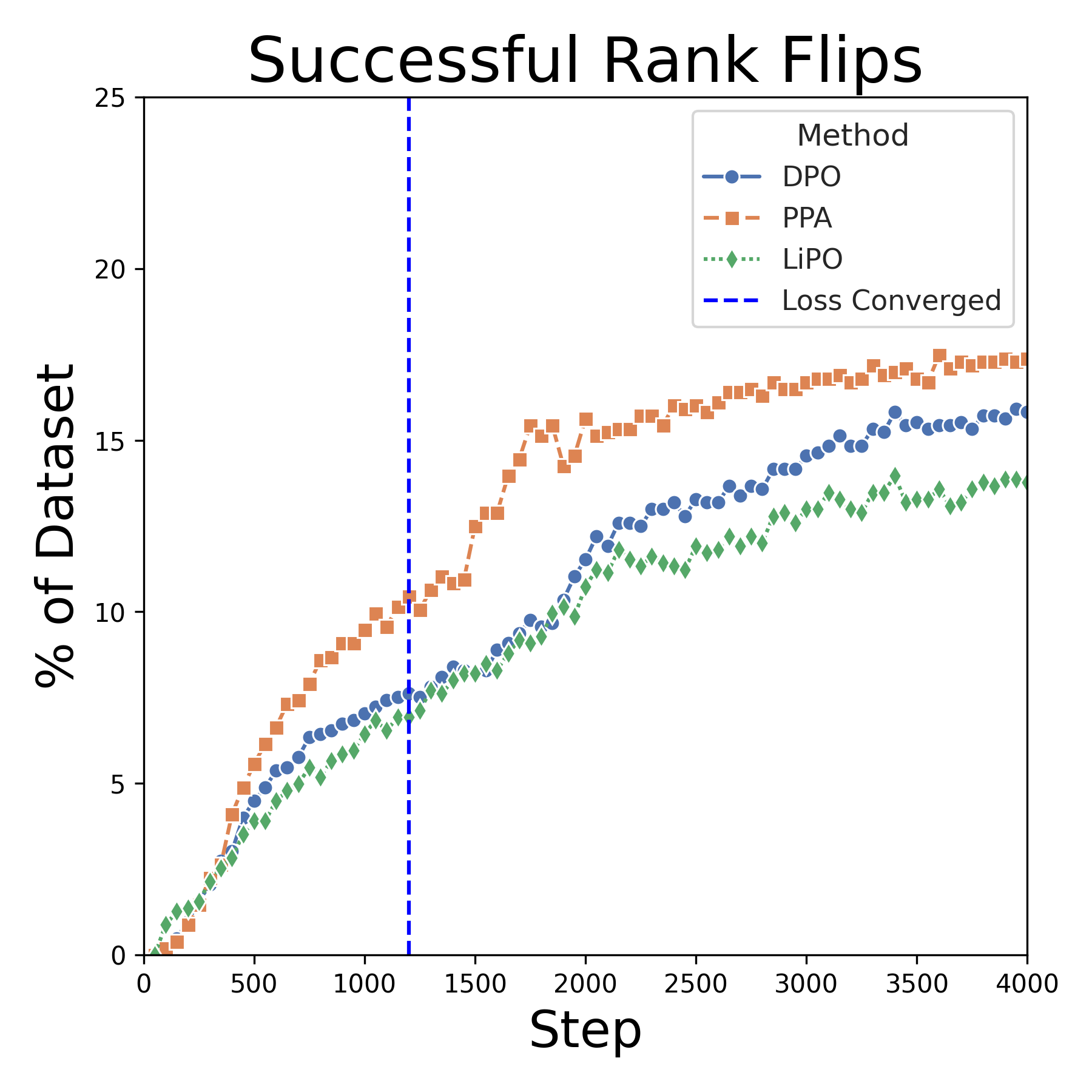} 
        \caption{Successful rank flip ratio.}
        \label{fig:subfig1}
    \end{subfigure}
    \begin{subfigure}[b]{0.48\textwidth}
        \centering
        \includegraphics[width=\textwidth]{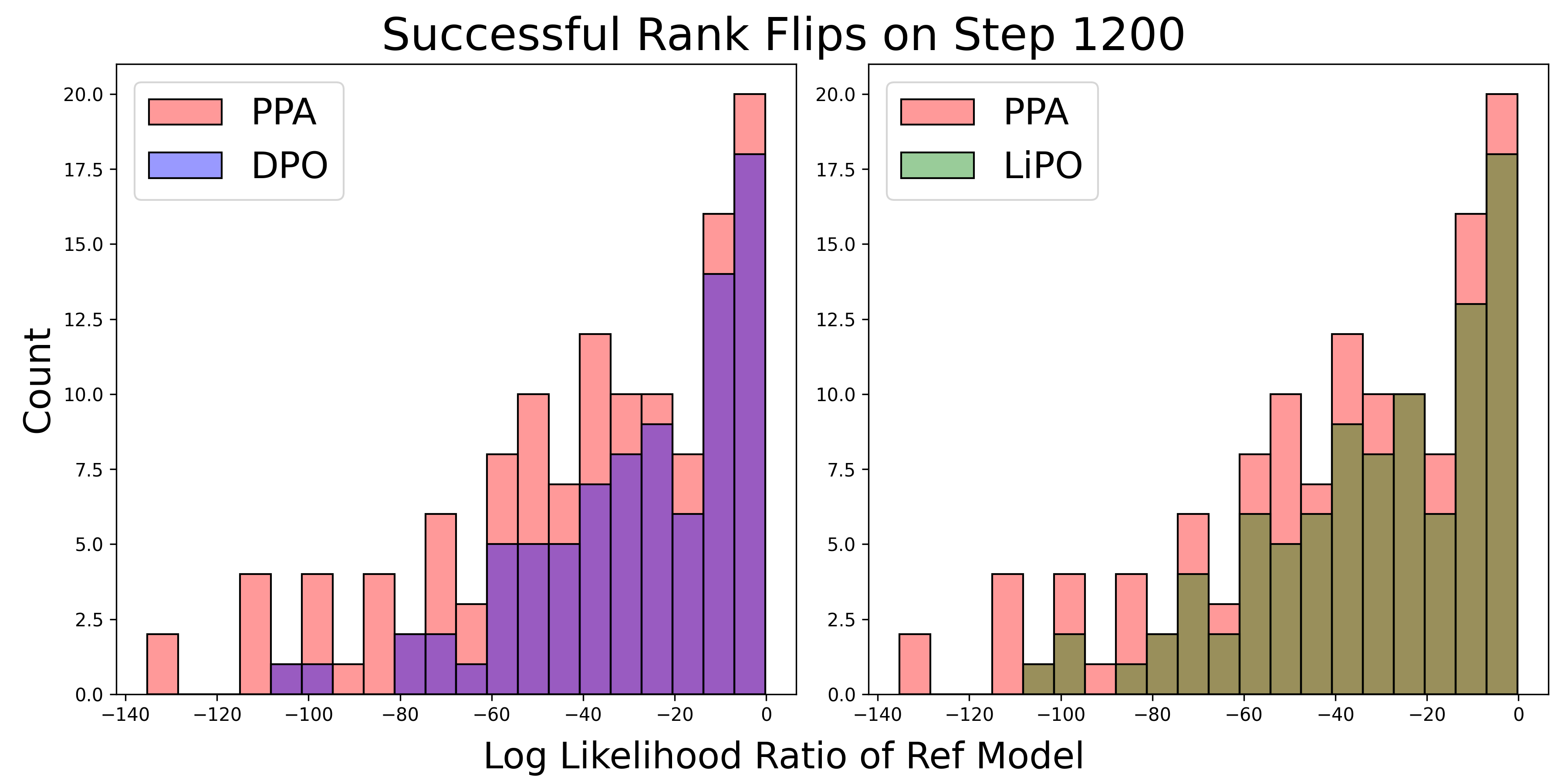} 
        \caption{Successful rank flip histogram.}
        \label{fig:subfig2}
    \end{subfigure}
    \begin{subfigure}[b]{0.24\textwidth}
        \centering
        \includegraphics[width=\textwidth]{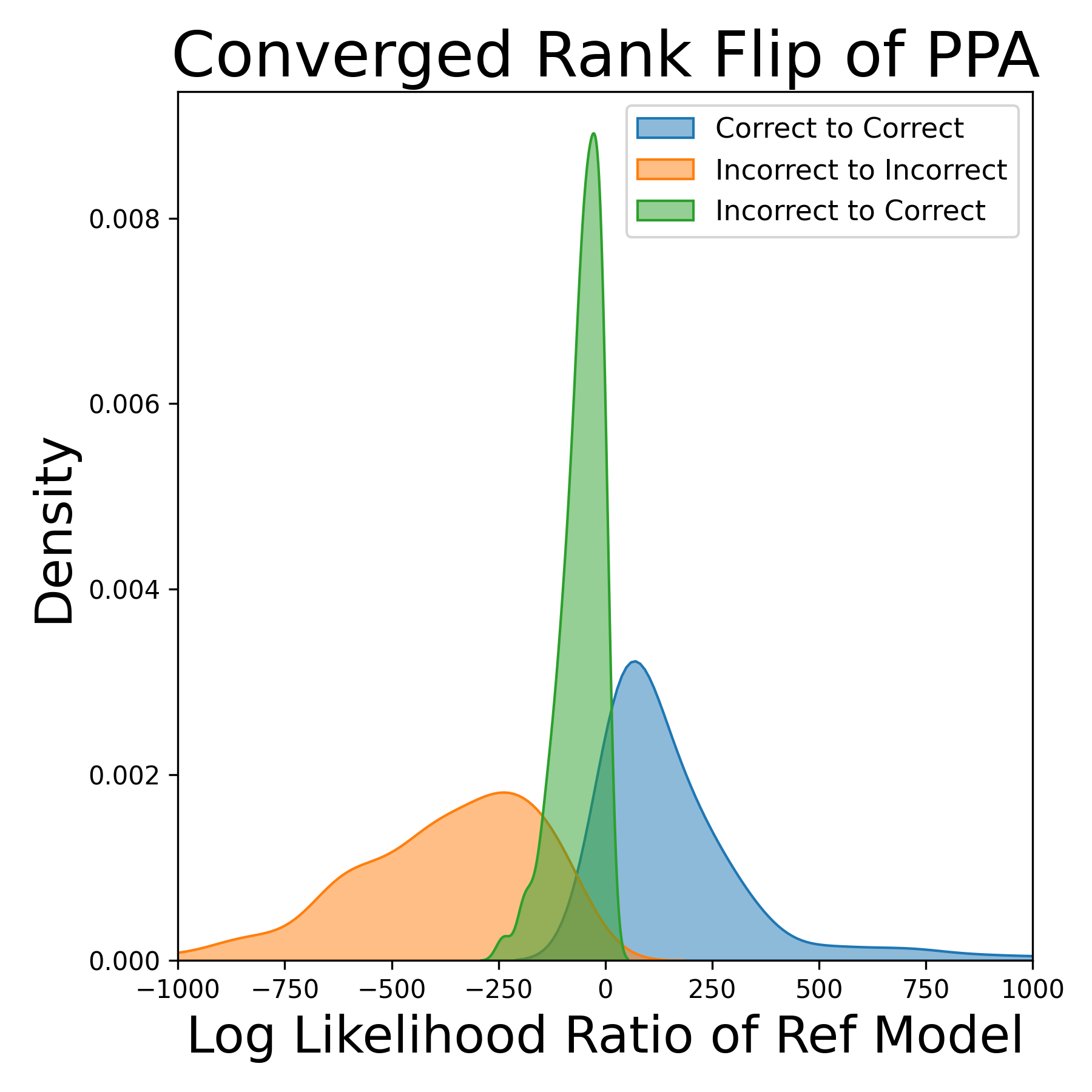} 
        \caption{Rank flip distribution.}
        \label{fig:subfig3}
    \end{subfigure}
    \caption{Comparisions among PPA, DPO, and LiPO on rank flips. (a) PPA demonstrates a higher efficiency in successful rank flips. The dashed line refers to the steps in which the loss objective is converged for all three methods. (b) PPA demonstrates more successful rank flips in loss-converged steps compared to DPO and LiPO. (c) The successful flip (Incorrect to Correct) distribution is highly constrained to reference ranking accuracy $y_w$ and $y_l$.}
    \label{fig:rank_flip}
\end{figure*}

\section{Permutative Preference Alignment}
In LLM alignment, the reward scores $\mathbf{s}$ in \Cref{scores_eq} are key to connecting the loss to the model parameters $\theta$. However, there is a gap between using NDCG as an evaluation metric and as a training objective. Since NDCG is non-differentiable with respect to reward scores $\mathbf{s}$, gradient descent cannot be directly applied for optimization. ~\looseness=-1

To overcome this limitation, surrogate losses \citep{ndcg_sur} have been developed. These losses approximate the NDCG value by converting its discrete and non-differentiable characteristics into a continuous and score-differentiable form, suitable for backpropagation. The original NDCG is computed by iterating over each list element's gain value and multiplying it by its corresponding position discount, a process known as the \emph{pairing between gains and discounts}. Thus, surrogate losses can be interpreted in two parts: pairing gains and discounts to approximate the NDCG value, and ensuring these functions are differentiable to enable gradient descent optimization. We will leverage NeuralNDCG \citep{neuralndcg} as such a surrogate loss.

\subsection{PPA Objective}
Our PPA incorporates a score-differentiable sorting algorithm-NeuralSort \citep{neuralsort}-to align gain values $G(\cdot)$ with position discounts $D(\cdot)$. This sorting operation is achieved by left-multiplying a permutation matrix $P_{\text{sort}(\mathbf{s})}$ in Eq \ref{neural_sort} with the score vector $\mathbf{s}$ to obtain a list of scores sorted in descending order. The element $P_{\text{sort}(\mathbf{s})}[i,j]$ denotes the probability that response $y_j$ is ranked in the $i$-th position after re-sorting based on $\mathbf{s}$. Applying this matrix to the gains $G(\cdot)$ results in the sorted gains vector $\widehat{G(\cdot)}$, which is aligned with discounts. Details about NeuralSort are shown in the Appendix \ref{appendix_neural_sort}. 
%including the specific simulation in \Cref{tab_p_hat} and code in \Cref{appendix_neural_sort_code}. 
For simplicity, we denote $\widehat{P}_{\text{sort}(\mathbf{s})}$ as $\widehat{P}$.

Similar to the original NDCG, but with the gain function $G(\cdot)$ replaced by $\widehat{G(\cdot)} = \widehat{P} \cdot G(\cdot)$ to ensure proper alignment between gains and discounts. The estimated gain at rank $j$ can be interpreted as a weighted sum of all gains, where the weights are given by the entries in the $j$-th row of $\widehat{P}$. Since $\widehat{P}$ is a row-stochastic matrix, each row sums to one, though the columns may not. This can cause $\widehat{G}$ to disproportionately influence the NDCG value at certain positions. To address this issue, we use the Sinkhorn scaling \citep{sinkhorn} on $\widehat{P}$ to ensure each column sums to one. Then we get:
\begin{equation}
\begin{aligned}
    \text{NeuralNDCG@}k\ (\tau;\mathbf{s}, \mathbf{\Psi}) = &\\
    &\hspace{-10em}N_k^{-1} \sum_{j=1}^{k} (\text{scale}(\widehat{P}) \cdot G(\mathbf{\Psi}))_j \cdot D(j)
\end{aligned}
\label{neural_ndcg}
\end{equation}
where $N_k^{-1}$ represents the $\text{maxDCG@}k$ (for $k \leq K$) as defined in Eq \ref{ndcg}. The function $\text{scale}(\cdot)$ denotes Sinkhorn scaling, and $G(\cdot)$ and $D(\cdot)$ are the gain and discount functions, respectively, as in Eq \ref{dcg}. The proposed PPA is illustrated in \Cref{fig:ill}. Intuitively, the gain function should be proportional to the label, effectively capturing the relative ranking of different responses. The discount function penalizes responses appearing later in the sequence, as in many generation or recommendation tasks, the focus is on the top-ranked elements. Thus, higher-ranked responses have a more significant impact on the overall loss in NeuralNDCG. Further illustrations are provided in Appendix \ref{appendix_sort}.

Finally, we derive the PPA objective:
\begin{equation}
\begin{aligned}
    &\mathcal{L}_{\text{PPA}}(\pi_\theta,k;\pi_{\text{ref}})=\\
    &\resizebox{0.45\textwidth}{!}{$-\mathbb{E}_{(x, \mathbf{Y}, \mathbf{\Psi}) \sim \mathcal{D}} \big[ \sum_{j=1}^{k} (\text{scale}(\widehat{P}) \cdot G(\mathbf{\Psi}))_j \cdot D(j)/N_k$}
    \big].
\end{aligned}
\label{loss_neural}
\end{equation}
%Note that setting $k=2$ with $K>2$ is not equivalent to having a list size of $K=2$. The former indicates a focus on the top-2 responses from the entire list, where a higher rank signifies superior response quality. Conversely, $K=2$ typically refers to a binary contrastive scenario, classifying responses as positive or negative samples and maximizing the likelihood of preferred response $y_w$ over non-preferred $y_l$. In high-quality response pairs, labeling one as negative may adversely impact the generation quality of LLMs. PPA provides a more comprehensive view of relative proximities within multiple responses. In this work, we set $k=K$ by default.

\subsection{Other Approximation of NDCG}
In addition to aligning gains and discounts, we can modify the discount function to be differentiable. ApproxNDCG \citep{approxndcg} is proposed as an approximation to the rank position in the NDCG equation (Eq \ref{dcg}) using the sigmoid function:
\begin{equation}
\small
\begin{aligned}
    \widehat{\tau(j)}&=1+\sum_{i\ne j}\frac{\exp{(-\alpha(s_j-s_i))}}{1+\exp{(-\alpha(s_j-s_i))}}\\
    &=1+\sum_{i\ne j}\sigma(\alpha(s_i-s_j))
\end{aligned}
\label{approx_rank}
\end{equation}
As observed, if $s_i \gg s_j$, the descending rank position of $y_j$ will increase by $1$. Note that the hyperparameter $\alpha$ controls the precision of the approximation. We then obtain the estimated $\widehat{\tau(j)}$ and subsequently the ApproxNDCG objective:
\begin{equation}
\begin{aligned}
&\mathcal{L}_{\text{ApproxNDCG@K}}(\pi_\theta;\pi_{\text{ref}})=\\
    &-\mathbb{E}_{(x, \mathbf{Y}, \mathbf{\Psi}) \sim \mathcal{D}} \big[ \sum_{j=1}^{k} G(\psi_j) \cdot D(\widehat{\tau(j)})/N_k
    \big].
\end{aligned}
\label{loss_approx}
\end{equation}

\begin{figure*}[ht]
    \centering
    \includegraphics[width=\textwidth]{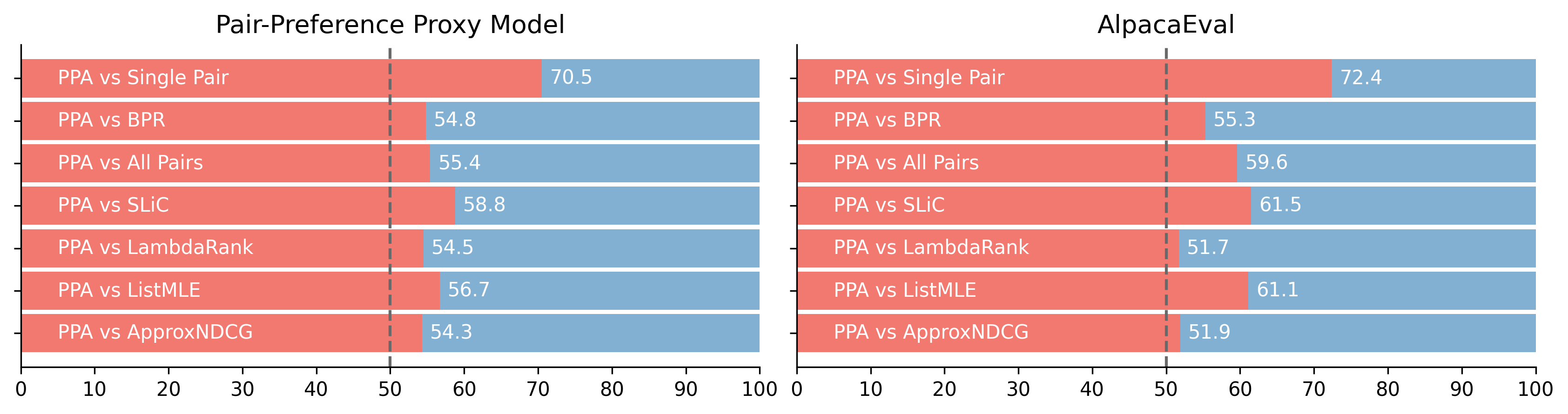}
    \caption{PPA outperforms other approaches on direct comparisons with Mistral-7B. The win rates are derived from comparisons between PPA and other methods on their optimal settings. We employ the Pair-Preference Proxy model on evaluation sets and GPT-4 on AlpacaEval as the judge models.}
    \label{fig_compare}
\end{figure*}

\section{Theoretical Analysis}
\subsection{Optimal Property}
\begin{property}
\label{prop:optimal}(Optimal property)
When \Cref{loss_neural} reaches the optimal value, the policy $\pi^*_{\theta,\text{NDCG}}$ is aligned with the reward model in terms of the response ranking permutations.
%The optimal policy from NDCG-based approaches $\pi^*_{\text{NDCG}}$ leads to the ground truth permutation from reward models.
\end{property}
The proof is shown in \Cref{proof:optimal}. NDCG achieves its maximum value if and only if the list permutation matches the ground-truth permutation. The distance between NDCG and its maximum value of 1 reflects the current alignment gap between the policy and the reward model.

\begin{proposition}
\label{prop:dpo_log2}
For DPO, in pairwise setting, correct ranking $s_w\geq s_l$ is achieved if and only if $\mathcal{L}_{\text{DPO}}\leq\log2$. But in listwise scenarios where $\text{list size}>2$, this condition on $\mathcal{L}_{\text{DPO}}$ no longer guarantees the correct overall ranking.
\end{proposition}
The proof is provided in \Cref{proof:dpo_log2}. This proposition indicates the limitations of DPO in handling multi-response scenarios. 
%In the training process of DPO, there is no intuitive bound that can reflect the alignment performance between the policy and the reward model. 
However, the NDCG-based method monitors the current alignment state and guarantees the correct list ranking. %, making it a more appropriate objective for handling multi-response scenarios.

%proof: s1 s2 s3 covariance

\subsection{Ranking Accuracy}
\begin{definition}
\label{def:rank_acc}(Ranking Accuracy \citep{rank_accuracy})
The ranking accuracy $\mathcal{R}$ of a model $\pi_\theta$ on a pairwise preference datapoint $(x,y_w,y_l)$ is 
\begin{equation*}
    \mathcal{R}(x, y_w, y_l; \pi_\theta) =
\begin{cases} 
1 & \pi_\theta(y_w \mid x) \geq \pi_\theta(y_l \mid x), \\ 
0 & \text{otherwise.}
\end{cases}
\end{equation*}

\end{definition}

\begin{definition}
\label{def:rank_flip}(Successful rank flip)
A successful rank flip is referred to as the model's ranking of responses shifts to favor the preferred over the non-preferred option:
$$
\pi_{\text{ref}}(y_w|x)\leq\pi_{\text{ref}}(y_l|x)\rightarrow \pi_\theta(y_w|x)\geq\pi_\theta(y_l|x)
$$
\end{definition}

\begin{proposition}
\label{prop:compare_ndcg_dpo}
    Assume log-likelihood ratio on reference model $X=\log\frac{\pi_{\text{ref}}(y_w|x)}{\pi_{\text{ref}}(y_l|x)}\sim N(0,\sigma^2)$ and after alignment training $Y=\log\frac{\pi_\theta(y_w|x)}{\pi_\theta(y_l|x)}\sim N(\mu_Y,\sigma_Y^2)$, we can get that the probability of successful rank flip of NDCG-based method is greater than DPO, which is $P_{\text{NDCG}}(Y>0|X<0)\geq P_{\text{DPO}}(Y>0|X<0)$.
\end{proposition}
The proof is provided in \Cref{proof:compare_ndcg_dpo}, which is also illustrated in \Cref{fig:rank_flip}. In alignment training, the objective is to increase the probability of the preferred response over the non-preferred one. A higher successful rank flip ratio indicates greater efficiency in achieving alignment.

\section{Experiments}

\begin{table*}[h]
\small
    \centering
    \begin{tabular}{lcccccc}
        \toprule
        \multirow{3}{*}{\textbf{\vspace{6pt}Method}}  & \multirow{3}{*}{\textbf{\vspace{6pt}Type}}& \multicolumn{2}{c}{\textbf{Proxy Model}} & \multicolumn{2}{c}{\textbf{General Benchmark}} & \multirow{3}{*}{\textbf{\vspace{6pt}Avg.}}\\
        \cmidrule(lr){3-4} \cmidrule(lr){5-6}
        &&  \textbf{Pair-Preference} &\textbf{Scoring} &   \textbf{AlpacaEval} & \textbf{MT-Bench} &\\
        \midrule
        Single Pair&Pairwise&60.75&56.86& 57.95& 52.81&\cellcolor{gray!20}57.09\\
        BPR & Pairwise & 60.32& 58.33 &58.74 &55.00&\cellcolor{gray!20}58.10\\
        All Pairs & Pairwise & 63.82& 60.54  & 57.23& 53.13&\cellcolor{gray!20}58.68\\
        RankNet & Pairwise & 62.27& 59.04  &58.94 &54.26 &\cellcolor{gray!20}58.63\\
        SLiC & Pairwise & 63.31& 60.70  & 61.00& 53.75&\cellcolor{gray!20}59.69\\
        LambdaRank & Listwise & 62.30& 59.04  & 58.72 &55.31&\cellcolor{gray!20}58.84\\
        ListMLE & Listwise & 63.03& 59.76  & 57.05 &53.13&\cellcolor{gray!20}58.24\\
        \midrule
        ApproxNDCG & Listwise & 61.46& 58.59  &58.16 &\underline{\textbf{55.94}}&\cellcolor{gray!20}59.33\\
        PPA & Listwise & \underline{\textbf{64.25}}& \underline{\textbf{61.36}}  & \underline{\textbf{61.64}} & 53.44&\cellcolor{gray!20}\underline{\textbf{60.17}}\\
        \bottomrule
    \end{tabular}
    \caption{The proposed PPA and ApproxNDCG outperform existing baselines across various evaluation benchmarks. The win rates are derived from comparisons between the preference-aligned Qwen2-0.5B and its SFT model. We set $\beta=0.1$ in Eq \ref{scores_eq} for all methods except $\beta=0.05$ for SLiC to achieve the optimal performance. 
}
    \label{tab: main results}
\end{table*}

\noindent\textbf{Baselines.}\quad  We employ various pairwise and listwise alignment baselines to explore the connection between LLM alignment and ranking tasks. Their optimization objectives are detailed in \Cref{baselines} of the appendix. We introduce three paradigms of positive-negative pairs for DPO on multiple responses. LiPO-$\lambda$ \citep{lipo} incorporates LambdaRank from the LTR literature, acting as a weighted version of DPO. SLiC and RRHF employ a similar hinge contrastive loss. ListMLE utilizes the Plackett-Luce Model \citep{PL1} to represent the likelihood of list permutations. For further information, please see Appendix \ref{appendix_baseline}.\\

\noindent\textbf{Datasets.}\quad We construct a multi-response dataset named \emph{ListUltraFeedback}\footnote{\url{https://huggingface.co/datasets/NDCG-alignment/ListUltraFeedback}}. This dataset combines four responses from UltraFeedback and five generated responses from the fine-tuned Llama3-8B model\footnote{\url{https://huggingface.co/datasets/princeton-nlp/llama3-ultrafeedback-armorm}} in SimPO \citep{cui2023ultrafeedback,simpo}, all based on the same prompts. All responses are assigned ground truth labels using the Reward Model ArmoRM \citep{ArmoRM}. This model is the leading open-source reward model, outperforming both GPT-4 Turbo and GPT-4o in RewardBench \citep{lambert2024rewardbench} at the time of our experiments. To ensure clear distinction between positive and negative samples, while maintaining diversity, we select two responses with the highest scores and two with the lowest. Additionally, we randomly draw four responses from the remaining pool. Details of the dataset are presented in \Cref{tab_datasets} of the appendix.\\

\noindent\textbf{Training Details.}\quad We use Qwen2-0.5B, Mistral-7B, and Llama3.1-8B \citep{qwen2,jiang2023mistral7b,llama3} as our foundation models, representing different parameter scales. Following the training pipeline in DPO, Zephyr, and SimPO, we start with supervised fine-tuning (SFT) on UltraChat-200k \citep{ding2023enhancing}. We then apply various pairwise and listwise approaches to align preferences on our multiple response dataset, ListUltraFeedback. Adhering to the settings in HuggingFace Alignment Handbook \citep{alignment_handbook2023}, we use a learning rate of $5 \times 10^{-7}$ and a total batch size of 128 for all training processes. The models are trained using the AdamW optimizer \citep{kingma2014adam} on 4 Nvidia V100-32G GPUs for Qwen2-0.5B models and 16 Nvidia V100-32G GPUs for Mistral-7B. Unless noted otherwise, we fix $\alpha=25$ for ApproxNDCG and $\tau=1$ for PPA to achieve optimal performance, as determined by ablation studies presented in Section \ref{ablation_study}. Both models and datasets are open-sourced, ensuring high transparency and ease of reproduction. Further training details can be found in Appendix \ref{appendix_train}.\\

\noindent\textbf{Evaluation.}\quad The KL-divergence in the original RLHF pipeline is designed to prevent the Policy model from diverging excessively from the SFT model, thus avoiding potential manipulation of the Reward Model. As we employ ArmoRM in the construction of the training dataset, we incorporate various judging models and evaluation benchmarks, such as different Reward models and AlpacaEval \citep{alpaca_eval} with GPT-4, to reduce the impact of overfitting on ArmoRM. We design 2 pipelines to thoroughly analyze the performance of PPA, using the Win Rate of policy models against SFT models as our primary metric. Details of evaluation datasets are presented in \Cref{tab_datasets} of the appendix.

In the \emph{Proxy Model} pipeline, we deploy the Scoring Reward Model ArmoRM\footnote{\url{https://huggingface.co/RLHFlow/ArmoRM-Llama3-8B-v0.1}} \citep{ArmoRM} and the Pair-Preference Reward Model\footnote{\url{https://huggingface.co/RLHFlow/pair-preference-model-LLaMA3-8B}} \citep{rlhflow} as Proxy Models to calculate the win rate on ListUltraFeedback. Both Proxy models surpass GPT-4 Turbo and GPT-4o in rewarding tasks on RewardBench \citep{lambert2024rewardbench}. The Scoring model provides a score in the range $(0,1)$ for a given prompt and response, while the Pair-Preference model outputs the winner when given a prompt and two responses, offering a more intuitive approach for pairwise comparisons.

In the \emph{General Benchmark} pipeline, we evaluate our models using two widely recognized benchmarks: AlpacaEval \citep{alpaca_eval} and MT-Bench \citep{mt-bench}, which assess the model's comprehensive conversational abilities across various questions. Consistent with the original setup, we employ GPT-4 Turbo \citep{gpt4} as the standard judge model to determine which of the two responses exhibits higher quality.

\subsection{Main Results}

\begin{figure*}[h]
    \centering
    \includegraphics[width=0.8\textwidth]{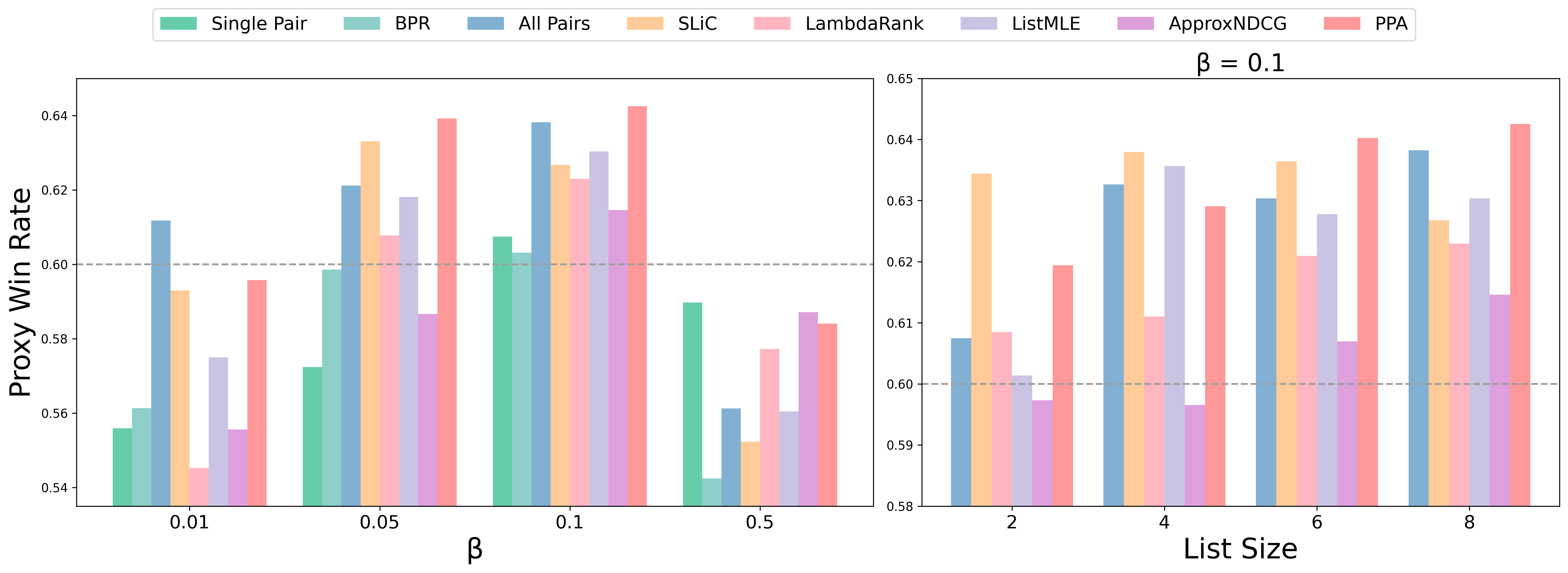}
    \caption{PPA outperforms other methods across different $\beta$ and list sizes. The Proxy win rates are calculated by Pair-Preference Proxy model by comparing preference-aligned Qwen2-0.5B against its SFT model.}
    \label{fig_combine}
\end{figure*}

\noindent\textbf{PPA significantly outperforms existing preference optimization baselines.}\quad We list win rates of various alignment approaches across diverse evaluation benchmarks in \Cref{fig_compare}, \Cref{tab: main results}, and \Cref{tab:llama3-8b}. Many approaches achieve their best performance with a list size of 8. As shown in \Cref{fig_combine}, PPA consistently outperforms other approaches when $K>4$, with performance improving as the list size increases. This trend is also evident across different values of $\beta$ in \Cref{tab_sup_list} of the appendix. PPA's advantage over pairwise and ListMLE methods lies in its efficiency in improving permutation performance. Traditional contrastive pairwise approaches maximize the likelihood of $y_w$ over $y_l$, which adversely affects the generation quality of LLMs when high-quality responses are treated as negative samples. In contrast, PPA provides a more holistic approach to handling the relationships between responses.\\

\noindent\textbf{PPA achieves higher ranking accuracy than DPO and LiPO.}\quad As illustrated in \Cref{fig:rank_flip}, PPA demonstrates a higher efficiency on successful rank flips. The difference in model performance is attributed to the distinction between the optimization objectives of the Bradley-Terry-based methods and NDCG-based methods. Furthermore, \Cref{fig:subfig3} reveals a significant correlation between the type of rank flip and the log-likelihood ratio of the reference model. Based on this observation, we hypothesize that the log-likelihood ratio of the policy model encodes the conditional probability of the log-likelihood ratio of the reference model.  \Cref{prop:compare_ndcg_dpo} provides a potential theoretical explanation for this improvement.\\

\noindent\textbf{List Size matters more than the number of pairwise comparisons.}\quad 
To assess the effect of varying response quantities and comparison methods, we conduct empirical studies across different list sizes (i.e., the number of responses the model can access) and comparison methods. Specifically, we test four pairwise comparison methods: Single Pair, BPR, OvW (i.e., Others vs Worst), and All Pairs in \Cref{baselines} of the appendix, which differ in the number of comparisons ($1$ comparison for Single Pair, $K-1$ for BPR and OvW, and $\binom{K}{2}$ for All Pairs).  \Cref{tab: main results} illustrates that both BPR and All Pairs methods outperform Single Pair, with no significant difference observed when $\text{List Size}=8$. This trend is particularly pronounced with the Mistral-7B model in \Cref{mistral7b} of the appendix. These findings suggest that the list size plays a more critical role than the specific number of pairwise comparisons.
%\Cref{fig_combine} indicates that models trained with multiple responses (more than two) significantly outperform those using binary responses. 
%\footnote{For the Single Pair approach, list sizes remain constant, as detailed in Section \ref{Single Pair}. In the case of BPR \citep{BPR}, since it focuses on the expected difference between the best response and others, list size has minimal impact in a random selection context.} 
\\

\subsection{Ablation Study}
\label{ablation_study}
\noindent\textbf{Score Function Scale.}\quad The parameter $\beta$ controls the scaling of the score function (Eq \ref{scores_eq}) and the deviation from the base reference policy $\pi_{\text{ref}}$, which significantly influences model performance. Following the existing work setting \citep{dpo,simpo, lipo}, we search $\beta$ among $[0.01, 0.05, 0.1, 0.5]$ and conduct sensitivity analysis. \Cref{fig_combine} shows that all methods achieve their best performance at $\beta=0.1$ except SLiC. PPA consistently achieves the best performance on both $\beta=0.05$ and $\beta=0.1$. Detailed results are in \Cref{tab_sup_beta} of the appendix.  \\

%\noindent\textbf{List Size.}\quad To evaluate the effectiveness of listwise methods in leveraging the sequential structure of multiple responses compared to pairwise methods, we analyze performance across varying list sizes.\footnote{For the Single Pair approach, list sizes remain constant, as detailed in Section \ref{Single Pair}. In the case of BPR \citep{BPR}, since it focuses on the expected difference between the best response and others, list size has minimal impact in a random selection context.} The results, presented in \Cref{fig_combine}, indicate that models trained with multiple responses (more than two) significantly outperform those using binary responses. Many models achieve optimal performance with a list size of 8. Notably, PPA Eq \ref{neural_ndcg} consistently outperforms other approaches when $K>4$, with performance improving as list size increases. This trend is also evident across different values of $\beta$ in \Cref{tab_sup_list}.\\

\begin{figure}[h]
\centering
\includegraphics[width=0.45\textwidth]{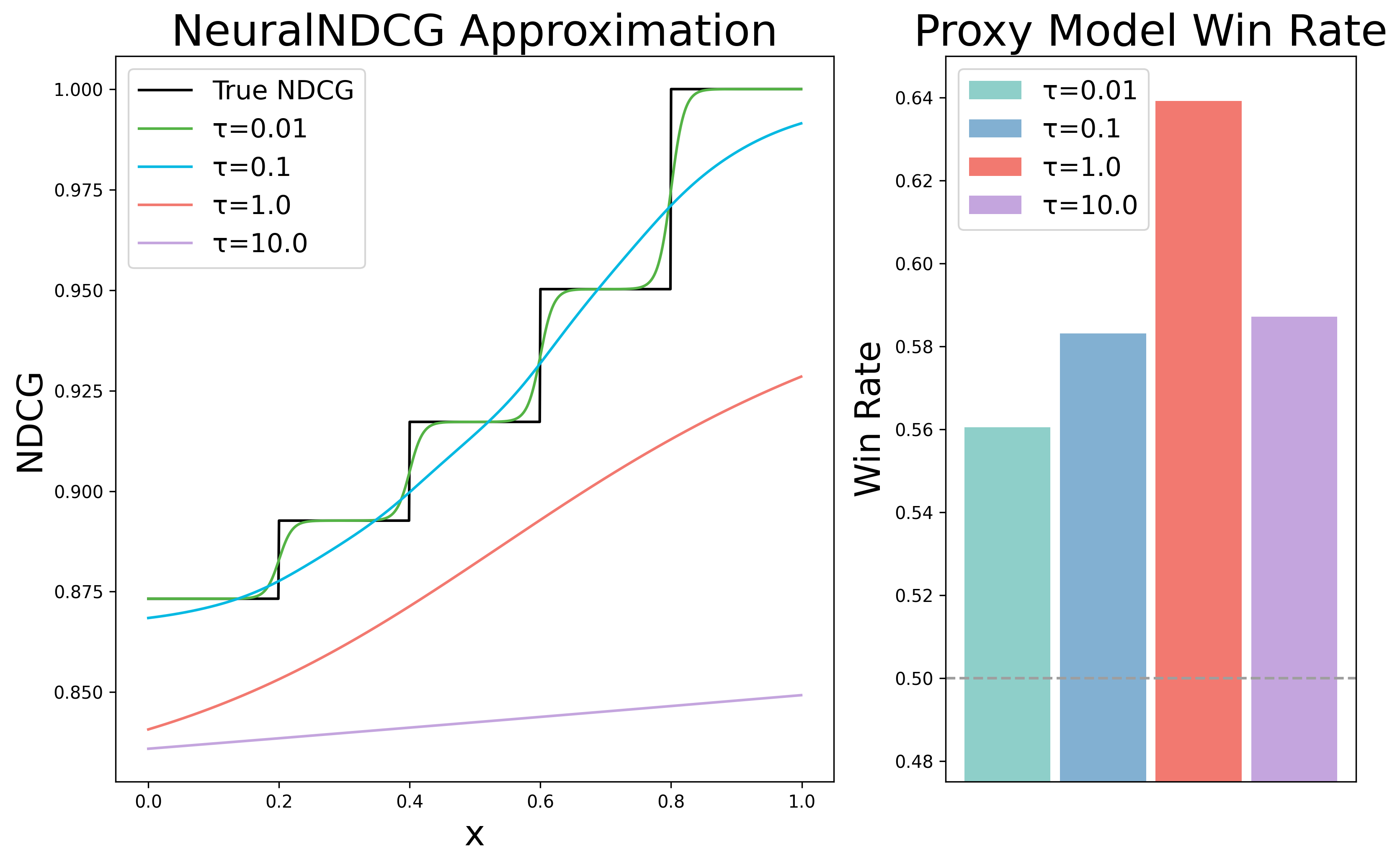}
\caption{Higher NDCG approximation accuracy does not always lead to better performance. Given ground truth label $\psi=[1.0,0.8,0.6,0.4,0.2]$ and  scores $\mathbf{s}=[x,0.8,0.6,0.4,0.2]$, an illustration of NeuralNDCG Approximation Accuracy with different $\tau$ and Pair-Preference Proxy win rates against SFT.}
\label{fig_neural}
\end{figure}

\noindent\textbf{Approximation Tradeoff.}\quad The temperature parameter $\tau$ controls the approximation accuracy and gradient variance of NeuralNDCG \citep{neuralndcg}. We visualize the values of NDCG and NeuralNDCG on specific data and assess model performance with various $\tau$. The results in \Cref{fig_neural} reveal that as NeuralNDCG more closely approximates true NDCG, model performance tends to decline. This may occur because training involves multiple high-quality responses with similar ground truth labels. Enforcing responses to conform to NDCG's step-wise structure can reduce the likelihood of good responses.

Additionally, as the approximation accuracy of NeuralNDCG increases, more plateaus appear due to NDCG's inherent step-wise nature. On these plateaus, gradients become zero, preventing model optimization. We visualize the loss landscape in \Cref{fig_ppa_gradient} in the Appendix and find that with low $\tau$, the gradients become highly spiky and discontinuous, making optimization challenging and potentially unstable. In contrast, higher $\tau$ values yield smoother and more navigable gradients. Further discussion is provided in Appendix \ref{appendix_approx_tradeoff}. A similar observation is confirmed on ApproxNDCG in Appendix \ref{appendix_approx}.\\

\noindent\textbf{PPA Setup.}\quad We perform an ablation study on key components of PPA.  As shown in \Cref{setup}, we find that (\romannumeral 1) when evaluating  NDCG@4 for multiple responses with a list size of 8, the performance is comparable to PPA with a list size of 4. This suggests that PPA's effectiveness is more influenced by the list size rather than the $k$ value in $\text{NDCG@}k$. (\romannumeral 2) The choice of gain function: $G_i = 2^{\psi_i} - 1$ or $G_i = \psi_i$, does not significantly impact model performance. The critical factor is that the gain provides the correct order of responses. (\romannumeral 3) Omitting Sinkhorn scaling \citep{sinkhorn} on $\widehat{P}$ significantly degrades performance. Without scaling, $\widehat{P}_{\text{sort}}$ may not be column-stochastic, meaning each column may not sum to one. Then the weighted sum of $G(\cdot)$ could disproportionately contribute to the estimated gain function $\widehat{G(\cdot)}$ and adversely affect performance.\\

\begin{table}[h]
\small
    \centering
    
    \begin{tabular}{lcc}
        \toprule
        \textbf{Method} & \textbf{Pair-Preference }& \textbf{Scoring } \\
        \midrule
        All Pairs     & 63.82   & 60.54 \\
        PPA    & \underline{\textbf{64.25}}  & \underline{\textbf{61.36}}\\
        
        \midrule
        Top-4    & 61.92  & 59.35 \\
        w/o Power  & 63.49& 61.28 \\
        w/o Scale   & 57.32& 56.20 \\
        \midrule
        $1/\tau$    & 62.78  & 59.76 \\
        $1/\sqrt{\tau}$  & 63.16& 60.19 \\
        $1/\tau^2$   & 62.47& 60.19 \\
        \bottomrule
    \end{tabular}
\caption{Ablation results for PPA Setup on Qwen2-0.5B: (Top) original setup; (Middle) ablation study on key components: k-value, gain function, and the Sinkorn Scale function; (Bottom) different discount settings.
%(\romannumeral 1) NDCG $k=4$; (\romannumeral 2) Use the direct label as the gain function; (\romannumeral 3) Remove the Sinkorn Scale function in Eq \ref{loss_neural};(\romannumeral 4)-(\romannumeral 6) different discount settings.
}
\label{setup}
\end{table}
\noindent\textbf{Model Scale Up}\quad To thoroughly assess the performance of PPA, we employ the Llama3.1-8B and Mistral-7B model as the LLM. Following the SimPO pipeline, we use their Instruct version as the SFT model. They are then aligned with multiple preferences on ListUltraFeedback, and the performance is validated across several benchmarks, as shown in \Cref{tab:llama3-8b}, \Cref{mistral7b}, and \Cref{app_tab:llama3-8b} of the appendix. Hyperparameter details and additional results are provided in Appendix \ref{appendix_train} and \ref{appendix_sup_mistral}.

PPA demonstrates competitive performance on win rates against the SFT model. To clearly illustrate PPA's advantages over other methods, we compare their generated responses and present PPA's win rates in \Cref{fig_compare,fig_sup_compare} of the appendix.\\

\begin{table}[h]
\small
    \centering
    \setlength{\tabcolsep}{4pt}
    \begin{tabular}{lcccc}
        \toprule
        \textbf{Method} & \textbf{Pair-Preference }& \textbf{Scoring }& \textbf{AlpacaEval} \\
        \midrule
        All Pairs     & 72.96   & 74.39&59.64& \\
        SLiC&72.84&75.04&60.20\\
        ListMLE &72.46&74.77&59.83\\
        LambdaRank&71.80&72.74&60.38\\
        \midrule
        PPA    & \underline{\textbf{74.34}}  & \underline{\textbf{75.58}}&\underline{\textbf{61.32	}}\\
        \bottomrule
    \end{tabular}
\caption{Model Scale Up: Our method PPA outperforms other approaches on Llama3.1-8B.
}
\label{tab:llama3-8b}
\end{table}

\noindent\textbf{Human Evaluation}\quad Human assessments provide crucial validation beyond proxy metrics. To address it, we conducted a comprehensive human evaluation involving 20 raters on the Prolific platform. Each rater compared 15 pairs of model outputs, with a total of 300 pairs evaluated across three matchups (PPA vs SFT, PPA vs DPO, PPA vs LiPO). The pairs were randomly selected to ensure an unbiased assessment.

\begin{table}[h]
\small
    \centering
    \setlength{\tabcolsep}{4pt}
    \begin{tabular}{lccc}
        \toprule
        \textbf{Comparison} & \textbf{PPA vs SFT}& \textbf{PPA vs DPO}& \textbf{PPA vs LiPO} \\
        \midrule
        Pair-Preference&70.5&55.4&54.5\\
        Scoring&68.9&55.2&57.0\\
        AlpacaEval&72.4&59.6&51.7\\
        \midrule
        HumanEval&68.0&55.0&54.0\\
        \bottomrule
    \end{tabular}
\caption{Human Evaluation result for comparisons. It shows PPA outperforms other baselines, which aligns with other automatic metrics.
}
\label{tab:human_eval}
\end{table}

The results in \Cref{tab:human_eval} demonstrate that PPA achieves consistent improvements over baselines like SFT, DPO, and LiPO, as confirmed by both human raters and automatic metrics. They also validate the consistency between the reward models and human judgments.

\section{Conclusion}
We propose Permutative Preference Alignment (PPA) to align listwise human judgments by optimizing the ranking metric NDCG. Empirical studies show that PPA consistently outperforms existing pairwise and listwise baselines across various setups. We identify the potential limitation of Bradley-Terry-based methods like DPO. PPA also demonstrates a higher efficiency in improving ranking accuracy and we propose a theoretical explanation for this improvement.

\section*{Limitations} 
Our study has several limitations and suggests promising directions for future research.
In constructing multiple responses, a pre-trained Reward Model serves as the judge model, which might not fully align with real-world human preferences. Future studies can develop more robust data construction methods to ensure responses remain harmless. Additionally, the extensive LTR literature remains underexplored, indicating potential for further research and applications in alignment fields.

\section*{Acknowledgements} 
YW was supported in part by the Office of Naval Research under grant N00014-23-1-2590, the National Science Foundation under grant No. 2231174, No. 2310831, No. 2428059, No. 2435696, No. 2440954, and a Michigan Institute for Data Science Propelling Original Data Science (PODS) grant. MY was supported in part by a Marketing Science Institute fund.

%\clearpage
\bibliography{reference}

\clearpage
\appendix
\section{Related Work}
\label{appendix_related_work}
\noindent\textbf{Pairwise Preference Optimization}\quad Direct Preference Optimization (DPO) \citep{dpo} removes the necessity for an explicit reward model within the RLHF framework by introducing a novel algorithm to compute reward scores for each response. Similar to RLHF, DPO uses the Bradley-Terry (BT) model \citep{bradley1952rank} to align binary human preferences in a contrastive manner. Subsequent research, including methods like IPO, KTO, RPO, SimPO, and others \citep{reject_sampling,cringe,ipo,rpo,kto,orpo,rdpo,simpo}, focus on refining the reward function and the BT model to enhance performance and simplify the process. Additionally, iterative methods are developed to align pairwise preferences with a dynamic reference model \citep{iter_nash,iter_reason,iter_sdpo,iter_self}. They classify preferred responses $y_w$ as positive samples and non-preferred responses $y_l$ as negative samples. They infer relative quality rankings of responses by maximizing the pairwise choice probability of $r(x,y_w)$ over $r(x,y_l)$. However, under multiple response scenarios, they focus on maximizing the average value of pairwise contrastive probability. It fails to guarantee a hard constraint that every individual pair is correctly aligned with ground truth labels. FocalPO \citep{focalpo} assigns greater weights to more informative ranking pairs, which shares similar insights with the NDCG metric.\\

%These contrastive techniques are influenced by the quality and quantity of negative samples. As indicated by the contrastive learning literature, the presence of hard negatives and large batch size is crucial \citep{contrastive}. Incorporating trivial negatives can lead to suboptimal results; hence, leveraging multiple-response data can expand the pool of candidate samples, reducing the likelihood of trivial negatives.\\

\noindent\textbf{Multiple Responses Alignment}\quad Recent research has introduced simple and efficient methods to align human preferences across multiple responses. These approaches expand candidate responses from various LLMs such as ChatGPT, Alpaca, and GPT-4, assigning rewards via reward models or human feedback. RRHF\citep{rrhf} employs the same hinge objective as SLiC \citep{slic} on multiple responses through pairwise comparisons. LiPO-$\lambda$ \citep{lipo} incorporates LambdaRank \citep{lambda1} where higher-quality responses against lower-quality ones receive greater weights, acting as a weighted version of DPO. However, when handling high-quality response pairs, incorrectly classifying one of them as the negative sample and minimizing its likelihood can adversely affect LLM generation quality. Listwise methods offer a more nuanced approach to handling relationships between responses. DPO-PL \citep{dpo} and PRO \citep{PRO} employ the same PL framework \citep{PL1} but differ in their reward functions. LIRE \citep{LIRE} calculates softmax probabilities with a consistent denominator and multiplies them by corresponding rewards, functioning as a pointwise algorithm since permutations do not alter loss values. Despite their potential, current listwise techniques are not yet state-of-the-art in the learning-to-rank (LTR) literature, indicating a need for further research.\\

\noindent\textbf{Learning to Rank (LTR)}\quad LTR involves a set of machine learning techniques widely applied in information retrieval, web search, and recommender systems \citep{liu2009learning, karatzoglou2013learning, BPR_rnn, li2024learning}. The goal is to train a ranking model by learning a scoring function $s = f(x, y)$ that assigns scores to elements for ranking purposes. The loss is computed by comparing the current permutation with the ground truth, which updates the model parameters $\theta$. Loss functions in LTR are generally categorized into three types: pointwise, pairwise, and listwise. Pointwise and pairwise methods convert the ranking task into classification problems, often overlooking the inherent structure of ordered data. Conversely, listwise approaches \citep{xia2008listwise} directly tackle the ranking problem by considering entire ranking lists as training instances. This approach fully exploits the relative proximities within multiple responses, enhancing the understanding of the ranking relationships.

\section{Proof of Theoretical Analysis}
\subsection{Proof of Property \ref{prop:optimal}}
\label{proof:optimal}

Assume access to a list of ground truth labels in descending order $\Psi=\{\psi_1,...,\psi_K\}$, where $\psi_i\geq\psi_j$ if $i< j$. Now we have a score vector $\mathbf{s}=\{s_1,...,s_k\}$, the descending rank position of $s_i$ is denoted by $$\tau(i)=1+\sum_{j=1}^k\mathbb{I}_{s_i<s_j}.$$
According to the definition of NDCG Eq \ref{ndcg}, the maximum NDCG value is achieved when $\tau(i)=i$, which is equivalent to $s^*_i\geq s^*_j$ if $i< j$. The permutation of $\mathbf{s^*}$ is the same as the permutation of ground truth labels $\Psi=\{\psi_1,...,\psi_K\}$, where $\psi_i\geq\psi_j$ if $i< j$. Then we can say the current policy model $\pi^*_{\text{NDCG}}$ is aligned with the reward model RM in terms of response permutations.

\subsection{Proof of Proposition \ref{prop:dpo_log2}}
\label{proof:dpo_log2}

Under pairwise $(x,y_w,y_l)$ scenario, 
$$
\mathcal{L}_{\text{DPO}}=-\log\sigma(s_w-s_l).
$$
When $s_w=s_l$, $\mathcal{L}_{\text{DPO}}=\log2$. As $\sigma(x)$ is an increasing function, it is easy to derive that $\mathcal{L}_{\text{DPO}}\leq\log2\Leftrightarrow s_w\geq s_l$.

But when $\text{list size}>2$, the condition of $\mathcal{L}_{\text{DPO}}\leq\log2$ no longer guarantees the correct overall ranking. Here we take a triplet datapoint $(x,y_1,y_2,y_3)$ where $\psi_1\geq\psi_2\geq\psi_3$ for example:
$$
\small
\begin{aligned}
    &\mathcal{L}_{\text{DPO}}(x,y_1,y_2,y_3|\pi_\theta)
    \\&=-\frac{1}{3}\times[\log\sigma(s_1-s_2)+\log\sigma(s_1-s_3)+\log\sigma(s_2-s_3)]
    \\&=-\frac{1}{3}\times[\log\frac{e^{s_1}}{e^{s_1}+e^{s_2}}+\log\frac{e^{s_1}}{e^{s_1}+e^{s_3}}+\log\frac{e^{s_2}}{e^{s_2}+e^{s_3}}]
\end{aligned}
$$

We assume $(s_1,s_2,s_3)=(0.7,0.5,0.6)$ and we can calculate the DPO loss:
$$
\small
\begin{aligned}
    \mathcal{L}_{\text{DPO}}&=-\frac{1}{3}\times[\log\sigma(0.2)+\log\sigma(0.1)+\log\sigma(-0.1)]\\&=0.662<\log2
\end{aligned}
$$
But obviously $s_1>s_3>s_2$ is not a correct list ranking. In this case, even if the DPO loss falls below a certain threshold, it does not guarantee an accurate list ranking. This limitation arises because the optimization process focuses on adjusting the scores $s_i$ and $s_j$ to increase the overall expected value but does not enforce hard constraints like $s_i>s_j$ for each individual pair. The DPO objective adjusts $s_i-s_j$ in a soft, overall sense but may not result in all differences being positive.

\subsection{Proof of Proposition \ref{prop:compare_ndcg_dpo}}
\label{proof:compare_ndcg_dpo}
We assume the log-likelihood ratios of all models follow normal distributions. Before alignment training, it is rational to hypothesize that the reference model cannot always distinguish and prioritize the preferred response $y_w$, so we set 0 to its mean:
$$
X=\log\frac{\pi_{\text{ref}}(y_w|x)}{\pi_{\text{ref}}(y_l|x)}\sim N(0,\sigma_X^2)
$$

Then after training, the distribution of log-likelihood ratio shifts on the policy model:
$$
Y=\log\frac{\pi_\theta(y_w|x)}{\pi_\theta(y_l|x)}\sim N(\mu_Y,\sigma_Y^2),
$$

As $X$ and $Y$ both follow normal distributions, we can use an independent variable $c$ from normal distributions to represent the \textit{training effect}. Since the parameters of independent normal distributions are additive, it is easier to write as follows
$$
Y=X+c,\ c\sim N(\mu_c,\tau_c^2),
$$
where $\mu_c=\mu_Y,\tau_c^2=\sigma_Y^2-\sigma_X^2$.

As $s_w=\beta\log\frac{\pi_\theta(y_w|x)}{\pi_{\text{ref}}(y_w|x)}$, we have
$$
\begin{aligned}
    c&=Y-X=\log\frac{\pi_\theta(y_w|x)}{\pi_\theta(y_l|x)}-\log\frac{\pi_{\text{ref}}(y_w|x)}{\pi_{\text{ref}}(y_l|x)}\\&=\log\frac{\pi_\theta(y_w|x)}{\pi_{\text{ref}}(y_w|x)}-\log\frac{\pi_\theta(y_l|x)}{\pi_{\text{ref}}(y_l|x)}=\frac{s_w-s_l}{\beta}
\end{aligned}
$$

Under pairwise scenarios, DPO-PL is equivalent to DPO-BT. The training objectives of DPO and NDCG are as follows:
\begin{equation}
\begin{aligned}
&\text{DPO}: \\&
\hspace{2em}\text{max } \mathbb{E}_{x,\mathbf{y},\psi\sim\mathcal{D}}(s_w-s_l)\equiv \text{max } E(c)\\
&\text{NDCG}: \\&
\hspace{2em}\text{max } \mathbb{E}_{x,\mathbf{y},\psi\sim\mathcal{D}}(\mathbb{I}_{s_w>s_l})\equiv \text{max } E(P(c>0))\\
\end{aligned}
\end{equation}

Based on the different training objectives of DPO and NDCG, we make the following assumptions. For NDCG method, the training objective is maximizing the number of cases where $s_w>s_l$, we assume a small mean $\mu_{\text{NDCG}}$ that is just enough to make $s_w$ likely to be greater than $s_l$. The variance $\tau_{\text{NDCG}}^2$ is also small, leading to a distribution concentrated around its mean. The assumptions above can result in a high probability of $P(c>0)$, which is aligned with NDCG's objective. During training, the algorithm adjust $s_w$ and $s_l$ to reduce the variability $\tau_{\text{NDCG}}^2$ of $c$, keeping differences small but positive.

Regarding DPO (same for LiPO and other pairwise contrastive methods), we assume a larger positive mean $\mu_{\text{DPO}}$ which increases the overall score margin. $c$ of DPO has a large variance $\tau_{\text{DPO}}^2$ as well, allowing for more variability in the differences. During training, the distributions of $s_w$ and $s_l$ are more spread out, leading to larger differences in $s_w-s_l$. The algorithm accepts higher variability and occasional negative differences to achieve a higher overall sum of $s_w-s_l$.

\Cref{fig:subfig3} indicates that log-likelihood ratio of the policy model is conditional on that of the reference model
then we have
$$
\begin{aligned}
&\text{NDCG}:\log\frac{\pi_\theta(y_w|x)}{\pi_\theta(y_l|x)}|\log\frac{\pi_{\text{ref}}(y_w|x)}{\pi_{\text{ref}}(y_l|x)}\\&\hspace{2em}\sim N(\log\frac{\pi_{\text{ref}}(y_w|x)}{\pi_{\text{ref}}(y_l|x)}+\mu_{\text{NDCG}},\sigma^2+\tau_{\text{NDCG}}^2)\\
&\text{DPO}:\log\frac{\pi_\theta(y_w|x)}{\pi_\theta(y_l|x)}|\log\frac{\pi_{\text{ref}}(y_w|x)}{\pi_{\text{ref}}(y_l|x)}\\&\hspace{2em}\sim N(\log\frac{\pi_{\text{ref}}(y_w|x)}{\pi_{\text{ref}}(y_l|x)}+\mu_{\text{DPO}},\sigma^2+\tau_{\text{DPO}}^2)
\end{aligned}
$$
where $\mu_{\text{NDCG}}\leq\mu_{\text{DPO}}$, $\tau_{\text{NDCG}}^2\leq\tau_{\text{DPO}}^2$, $P(c_{\text{NDCG}}>0)\geq P(c_{\text{DPO}}>0)$.\\

As $c\sim N(\mu_c,\tau_c^2)$, we have $z=(c-\mu_c)/\tau_c \sim N(0,1)$, then we can derive
$$
P(c>0)=P(z>-\frac{\mu_c}{\tau_c})=P(z<\frac{\mu_c}{\tau_c})=\Phi(\frac{\mu_c}{\tau_c}),
$$
where $\Phi(\cdot)$ is the standard normal cumulative distribution function (CDF), which is an increasing function. Under assumptions above $P(c_{\text{NDCG}}>0)\geq P(c_{\text{DPO}}>0)$, we have 
\begin{equation}
    \frac{\mu_{\text{NDCG}}}{\tau_{\text{NDCG}}} \geq \frac{\mu_{\text{DPO}}}{\tau_{\text{DPO}}}
\label{eq_mutau_ratio}
\end{equation}
By definition, we refer to turning an incorrect pair from the reference model (i.e., $\frac{\pi_{\text{ref}}(y_w|x)}{\pi_{\text{ref}}(y_l|x)}<1$) into a correct pair in the policy model (i.e., $\frac{\pi_\theta(y_w|x)}{\pi_\theta(y_l|x)}>1$) as a successful rank flip. Then we can represent the probability of it via the annotations above:
$$
\begin{aligned}
    &P(\log\frac{\pi_\theta(y_w|x)}{\pi_\theta(y_l|x)}>0|\log\frac{\pi_{\text{ref}}(y_w|x)}{\pi_{\text{ref}}(y_l|x)}<0):=\\
    &P(Y>0|X<0)=\frac{P(Y>0,X<0)}{P(X<0)}.
\end{aligned}
$$
Since $X\sim N(0,\sigma^2)$, $P(X<0)=0.5$, so:
$$
\begin{aligned}
    &P(Y>0|X<0)=2P(Y>0,X<0)\\&\hspace{1em}=2\int_{x=-\infty}^0 P(Y>0|X=x)f_X(x)\ dx.
\end{aligned}
$$
Given $X=x, Y=x+c$,
$$
\begin{aligned}
&P(Y>0|X=x)=P(c>-x)\\
&\hspace{1em}=P(z>\frac{-x-\mu_c}{\tau_c})=\Phi(\frac{x+\mu_c}{\tau_c}).
\end{aligned}
$$

As $X\sim N(0,\sigma^2)$, we replace $x$ with parameter $s=x/\sigma\sim N(0,1)$, then we have:
\begin{equation}
    P(Y>0|X<0)=2\int_{s=-\infty}^0 \Phi(\frac{\sigma s+\mu_c}{\tau_c})\phi(s)\ ds,
\end{equation}

where $\phi(s)$ is the PDF of the standard normal distribution and $\Phi(\cdot)$ is the standard normal's CDF. Since $\Phi(\cdot)$ is an increasing function and in Eq \ref{eq_mutau_ratio} we have $\mu_{\text{NDCG}}/\tau_{\text{NDCG}}\geq\mu_{\text{DPO}}/\tau_{\text{DPO}}$ and $\tau_{\text{NDCG}}\leq\tau_{\text{DPO}}$, we have
$$
\Phi(\frac{\sigma s+\mu_{\text{NDCG}}}{\tau_{\text{NDCG}}})\geq \Phi(\frac{\sigma s+\mu_{\text{DPO}}}{\tau_{\text{DPO}}}),
$$

Finally we can derive:
$$
P_{\text{NDCG}}(Y>0|X<0)\geq P_{\text{DPO}}(Y>0|X<0)
$$

\section{Details of Baselines}
\label{appendix_baseline}
\Cref{baselines} shows the types and objectives of the baselines we consider in the empirical study.

\begin{table*}[t]
\small
\centering
\begin{tabular}{@{} l c l @{}}
\toprule
\textbf{Method} & \textbf{Type} &\textbf{Objective} \\
\midrule
DPO - Single Pair (\ref{Single Pair}) &Pairwise& $-\log \sigma \left( s_1-s_K \right)$ \\
\midrule
DPO - BPR (\ref{BPR})&Pairwise& $-\frac{1}{K-1} \sum_{j\ne 1}^K\log \sigma \left( s_1-s_j \right)$ \\
\midrule
DPO - Others vs Worst (\ref{ovw})&Pairwise& $-\frac{1}{K-1} \sum_{j\ne K}^K\log \sigma \left( s_j-s_K \right)$ \\
\midrule
DPO - All Pairs (\ref{all pairs}) &Pairwise&$-\binom{K}{2}^{-1} \sum_{\psi_i>\psi_j}\log \sigma \left( s_i-s_j \right)$\\
\midrule
SLiC (\ref{slic})&Pairwise&$-\binom{K}{2}^{-1}\sum_{\psi_i > \psi_j} \max(0,1-(s_i-s_j))$\\
\midrule
LambdaRank (\ref{lipo})&Listwise&$-\binom{K}{2}^{-1}\sum_{\psi_i > \psi_j} \Delta_{i,j} \log \sigma \left( s_i-s_j \right)$\\
&& where $\Delta_{i,j} = |G_i - G_j| \cdot \left| {D(\tau(i))} - {D(\tau(j))} \right|$\\
\midrule
ListMLE (\ref{listmle})&Listwise&$-\log\prod_{k=1}^K \frac{exp(s_k)}{\sum_{j=k}^K exp(s_j)}$\\
\midrule
\midrule
ApproxNDCG (\ref{loss_approx}) &Listwise&$-N_k^{-1} \sum_{j=1}^{k} G(\psi_j) \cdot D(\widehat{\tau(j)})$\\
\midrule
PPA (\ref{loss_neural}) &Listwise&$-N_k^{-1} \sum_{j=1}^{k} (\text{scale}(\widehat{P}) \cdot G(\mathbf{\Psi}))_j \cdot D(j)$\\
\bottomrule
\end{tabular}%
\caption{Pairwise and listwise baselines given multiple-response data $\mathcal{D} = (x, \mathbf{Y}, \mathbf{\Psi})$.}
\label{baselines}
\end{table*}

To ensure variable consistency and comparability of experiments, we choose the original DPO algorithm as our reward score function Eq \ref{scores_eq} and pairwise baseline method and assess its performance in both binary-response and multi-response scenarios.\\

\noindent\textbf{DPO-BT}\quad In detail, we implement four variants of the original sigmoid-based pairwise DPO based on the Bradley-Terry (BT) methods while aligning multiple responses. The first one is \textbf{Single Pair} paradigm, where we compare only the highest-scoring and lowest-scoring responses, which is equivalent to the original DPO in the pairwise dataset scenario.
\begin{equation}
\begin{aligned}
    \mathcal{L}_{\text{Single Pair}}(\pi_\theta;\pi_{\text{ref}})=&\\
    &\hspace{-6em}-\mathbb{E}_{(x, \mathbf{Y}, \mathbf{\Psi}) \sim \mathcal{D}} \left[ \log \sigma \left( s_1-s_K \right) \right],
\end{aligned}
\label{Single Pair}
\end{equation}

Then we introduce the \textbf{Bayesian Personalized Ranking (BPR)} \citep{BPR} algorithm that computes the response with the highest score against all other negative responses based on Bayes' theorem\footnote{The \textbf{BPR} variant Eq \ref{BPR} can be viewed as the expected loss function in the following scenario: we have a multiple responses dataset, but we only retain the highest-scoring response and randomly select one from the remaining. Finally, we construct a binary responses dataset for pairwise preference optimization, which is a widely used method for building pairwise datasets \citep{alignment_handbook2023,simpo}.}, which is widely used in recommender system \citep{BPR_rnn}.

\begin{equation}
\begin{aligned}
    \mathcal{L}_{\text{BPR}}(\pi_\theta;\pi_{\text{ref}})= &\\
    &\hspace{-6em}-\mathbb{E}_{(x, \mathbf{Y}, \mathbf{\Psi}) \sim \mathcal{D}} \left[\frac{1}{K-1} \sum_{j\ne 1}^K\log \sigma \left( s_1-s_j \right) \right],
\end{aligned}
\label{BPR}
\end{equation}
The third one is \textbf{Others vs Worst}, which calculates other responses against the worst one:
\begin{equation}
\begin{aligned}
    \mathcal{L}_{\text{OvW}}(\pi_\theta;\pi_{\text{ref}})= &\\
    &\hspace{-6em}-\mathbb{E}_{(x, \mathbf{Y}, \mathbf{\Psi}) \sim \mathcal{D}} \left[\frac{1}{K-1} \sum_{j\ne K}^K\log \sigma \left( s_j-s_K \right) \right],
\end{aligned}
\label{ovw}
\end{equation}
In the last BT variant, we consider all pairs that can be formed from K responses, which is similar to PRO \citep{PRO}. This approach allows the model to gain more comprehensive information than the aforementioned methods, including preference differences among intermediate responses, which is referred to as \textbf{All Pairs}:
\begin{equation}
\begin{aligned}
    \mathcal{L}_{\text{All Pairs}}(\pi_\theta;\pi_{\text{ref}})= &\\
    &\hspace{-7em}-\mathbb{E}_{(x, \mathbf{Y}, \mathbf{\Psi}) \sim \mathcal{D}} \left[\frac{1}{\binom{K}{2}} \sum_{\psi_i>\psi_j}\log \sigma \left( s_i-s_j \right) \right],
\end{aligned}
\label{all pairs}
\end{equation}
where $\binom{K}{2}$ denotes the number of combinations choosing 2 out of K elements.\\

\noindent\textbf{LiPO-$\lambda$}\quad Deriving from the LambdaRank \citep{lambda1}, the objective of LiPO-$\lambda$ \citep{lipo} can be written as follows:
\begin{equation}
\begin{aligned}
    \mathcal{L}_{\text{LambdaRank}}(\pi_\theta; \pi_{\text{ref}}, \beta) = &\\
    &\hspace{-11em} -\mathbb{E}_{(x, \mathbf{Y}, \mathbf{\Psi}) \sim \mathcal{D}} \left[\frac{1}{\binom{K}{2}}
    \sum_{\psi_i > \psi_j} \Delta_{i,j} \log \sigma \left( s_i-s_j \right)
    \right],
\end{aligned}
\label{lipo}
\end{equation}
$$
\text{where} \quad \small{\Delta_{i,j} = |G_i - G_j| \cdot \left| \frac{1}{D(\tau(i))} - \frac{1}{D(\tau(j))} \right|}.
$$
\(\Delta_{i,j}\) is referred to as the Lambda weight. \(G\) is known as a gain function, with \(G_i = 2^{\psi_i} - 1\) being a commonly used example. The function \(D\) serves as a discount function, with \(D(\tau(i)) = \log_2(1 + \tau(i))\), where \(\tau(i)\) is the same as in Eq \ref{dcg}.\\

\noindent\textbf{SLiC}\quad Following the analogous objectives proposed in RRHF \citep{rrhf} and SLiC \citep{slic}, we integrate the pairwise ReLU-based loss as one of our baselines:
\begin{equation}
\small
\begin{aligned}
    \mathcal{L}_{\text{SLiC}}(\pi_\theta;\pi_{\text{ref}})=&\\
    &\hspace{-7em}\mathbb{E}_{(x, \mathbf{Y}, \mathbf{\Psi}) \sim \mathcal{D}} \left[\frac{1}{\binom{K}{2}}\sum_{\psi_i > \psi_j} \max(0,1-(s_i-s_j)) \right],
\end{aligned}
\label{slic}
\end{equation}

\noindent\textbf{DPO-PL}\quad The DPO objective can also be derived under the Plackett-Luce Model \citep{PL1} in a listwise manner, which is equivalent to the ListMLE \citep{listmle} method:
\begin{equation}
\begin{aligned}
    \mathcal{L}_{\text{ListMLE}}(\pi_\theta;\pi_{\text{ref}})=&\\
    &\hspace{-8em} -\mathbb{E}_{(x, \mathbf{Y}, \mathbf{\Psi}) \sim \mathcal{D}} \left[ \log\prod_{k=1}^K \frac{exp(s_k)}{\sum_{j=k}^K exp(s_j)} \right],
\end{aligned}
\label{listmle}
\end{equation}

\section{NeuralSort Details}
\label{appendix_neural_sort}
\subsection{NeuralSort relaxation}
To approximate the sorting operator, we need to approximate this permutation matrix. In NeuralSort \citep{neuralsort}, the permutation matrix is approximated using a unimodal row stochastic matrix $\widehat{P}_{\text{sort}(\mathbf{s})}(\tau)$, defined as:
\begin{equation}
\widehat{P}_{\text{sort}(\mathbf{s})}[i, :](\tau) = \text{softmax}\left[\frac{((n + 1 - 2i)\mathbf{s} - A_{\mathbf{s}} \mathbf{1})}{\tau}\right].
\label{neural_sort}
\end{equation}
Here, $A_{\mathbf{s}}$ is the matrix of absolute pairwise differences of elements in $\mathbf{s}$, where $A_{\mathbf{s}}[i, j] = |s_i - s_j|$, and $\mathbf{1}$ is a column vector of ones. The row of $\widehat{P}_{\text{sort}(\mathbf{s})}$ always sums to one. The temperature parameter $\tau > 0$ controls the accuracy of the approximation.
%$\tau$ balances the trade-off between approximation accuracy and gradient variance. 
Lower values of $\tau$ yield better approximations but increase gradient variance:
\begin{equation}
\lim_{\tau \to 0} \widehat{P}_{\text{sort}(\mathbf{s})}(\tau) = P_{\text{sort}(\mathbf{s})}.
\label{neural_sort2}
\end{equation}
A specific simulation is shown in \Cref{tab_p_hat} of the appendix. For simplicity, we denote $\widehat{P}_{\text{sort}(\mathbf{s})}$ as $\widehat{P}$.

\subsection{NDCG Approximation}
\label{appendix_sort}
Given the input ground truth labels $\mathbf{\Psi}=[5,4,3,2]^T$ and scores $\mathbf{s} = [9, 1, 5, 2]^T$, the descending order of $\mathbf{\Psi}$ based on the current reward scores $\mathbf{s}$ is $\mathbf{\tau}=[1,4,2,3]^T$. According to the formula introduced in Eq \ref{dcg}:
$$
\small
\begin{aligned}
    \text{DCG@4}=\sum_{j=1}^{k} G(\psi_j) \cdot D(\tau(j))=&\\
    &\hspace{-15em} \frac{G(5)}{\log_2(1+1)}+\frac{G(4)}{\log_2(1+4)}+\frac{G(3)}{\log_2(1+2)}+\frac{G(2)}{\log_2(1+3)}
\end{aligned}
$$
Building upon the preliminaries defined in \citep{neuralsort}, consider an $n$-dimensional permutation $\mathbf{z} = [z_1, z_2, \ldots, z_n]^T$, which is a list of unique indices from the set ${1, 2, \ldots, n}$. Each permutation $\mathbf{z}$ has a corresponding permutation matrix $P_{\mathbf{z}} \in {0, 1}^{n \times n}$, with entries defined as follows:
\begin{equation}
P_{\mathbf{z}}[i,j] =
\begin{cases}
1 & \text{if } j = z_i \\
0 & \text{otherwise}.
\end{cases}
\end{equation}
Let $\mathbb{Z}_n$ denote the set containing all $n!$ possible permutations within the symmetric group. We define the $\text{sort}: \mathbb{R}^n \rightarrow \mathbb{Z}_n$ operator as a function that maps $n$ real-valued inputs to a permutation representing these inputs in descending order. 

The $\text{sort}(\mathbf{s}) = [1, 3, 4, 2]^T$ since the largest element is at the first index, the second largest element is at the third index, and so on. We can obtain the sorted vector simply via $P_{\text{sort}(\mathbf{s})} \cdot\mathbf{s}$:
\begin{equation}
P_{\text{sort}(\mathbf{s})}\cdot\mathbf{s} = 
\begin{pmatrix}
1 & 0 & 0 & 0 \\
0 & 0 & 1 & 0 \\
0 & 0 & 0 & 1 \\
0 & 1 & 0 & 0 \\
\end{pmatrix}
\begin{pmatrix}
9 \\
1 \\
5 \\
2 \\
\end{pmatrix}
=
\begin{pmatrix}
9 \\
5 \\
2 \\
1 \\
\end{pmatrix}
\label{psort}
\end{equation}
Here we demonstrate the results by conducting NeuralSort Relaxation Eq \ref{neural_sort} with different $\tau$. 
\begin{table}[h]
\centering
\resizebox{0.48\textwidth}{!}{%
\begin{tabular}{lcccc}
\toprule
& \multicolumn{4}{c}{$\widehat{P_{\text{sort}(\mathbf{s})}} \cdot \mathbf{s}$} \\
\midrule
$\lim_{\tau \to 0}$ & 9 & 5 & 2 & 1 \\
\midrule
$\tau=0.01$ & 9.0000 & 5.0000 & 2.0000 & 1.0000 \\ 
$\tau=0.1$ & 9.0000 & 5.0000 & 2.0000 & 1.0000 \\ 
$\tau=1.0$ & 8.9282 & 4.9420 & 1.8604 & 1.2643 \\
$\tau=10.0$ & 6.6862 & 4.8452 & 3.2129 & 2.2557 \\
\bottomrule
\end{tabular}
}
\caption{Illustration of Sorting Operation of ground truth labels $\mathbf{\Psi}=[5,4,3,2]^T$ and scores $\mathbf{s} = [9, 1, 5, 2]^T$ via NeuralSort \citep{neuralsort} with different $\tau$.}
\label{tab_p_hat}
\end{table}
When we integrate the NeuralNDCG formula in Eq \ref{neural_ndcg}, ideally, $\lim_{\tau \to 0} \widehat{P}_{\text{sort}(\mathbf{s})}(\tau) = P_{\text{sort}(\mathbf{s})}$, yielding the following result:
\[
\begin{aligned}
\widehat{G}=P_{\text{sort}(\mathbf{s})}\cdot\text{G}(\mathbf{\Psi}) =&\\ 
&\hspace{-7em}\begin{pmatrix}
1 & 0 & 0 & 0 \\
0 & 0 & 1 & 0 \\
0 & 0 & 0 & 1 \\
0 & 1 & 0 & 0 \\
\end{pmatrix}
\begin{pmatrix}
\text{G(5)} \\
\text{G(4)} \\
\text{G(3)} \\
\text{G(2)} \\
\end{pmatrix}
=
\begin{pmatrix}
\text{G(5)} \\
\text{G(3)} \\
\text{G(2)} \\
\text{G(4)} \\
\end{pmatrix}
\end{aligned}
\]
Then,
$$
\small
\begin{aligned}
    \text{NeuralDCG@4}= \sum_{j=1}^{k} (\widehat{G})_j \cdot D(j)=&\\ 
&\hspace{-15em}\frac{G(5)}{\log_2(1+1)}+\frac{G(3)}{\log_2(1+2)}+\frac{G(2)}{\log_2(1+3)}+\frac{G(4)}{\log_2(1+4)}
\end{aligned}
$$
which can be easily seen to be the same as DCG@4.

\subsection{Approximation Tradeoff Analysis}
\label{appendix_approx_tradeoff}

\begin{figure*}[h]
    \centering
    \includegraphics[width=\textwidth]{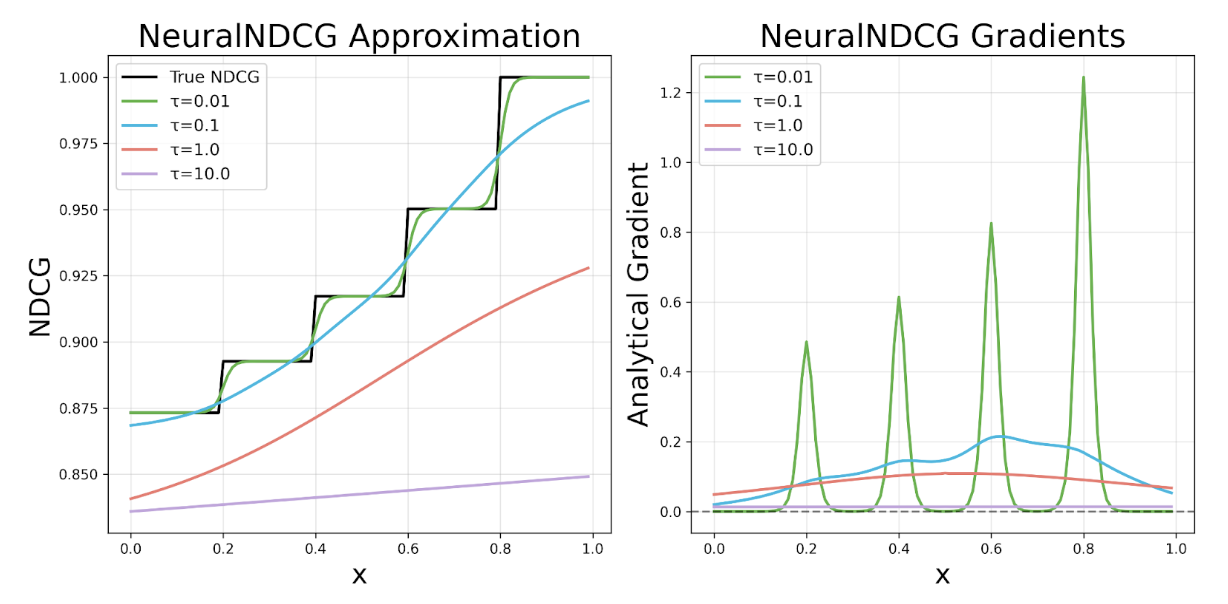}
    \caption{Given ground truth label $\psi=[1.0,0.8,0.6,0.4,0.2]$, the scores $\mathbf{s}=[x,0.8,0.6,0.4,0.2]$ and fix $\beta=0.1$, we visualize the PPA Approximation Accuracy with different $\tau$ and its corresponding gradients.}
    \label{fig_ppa_gradient}
\end{figure*}

\begin{figure*}[h]
    \centering
    \includegraphics[width=\textwidth]{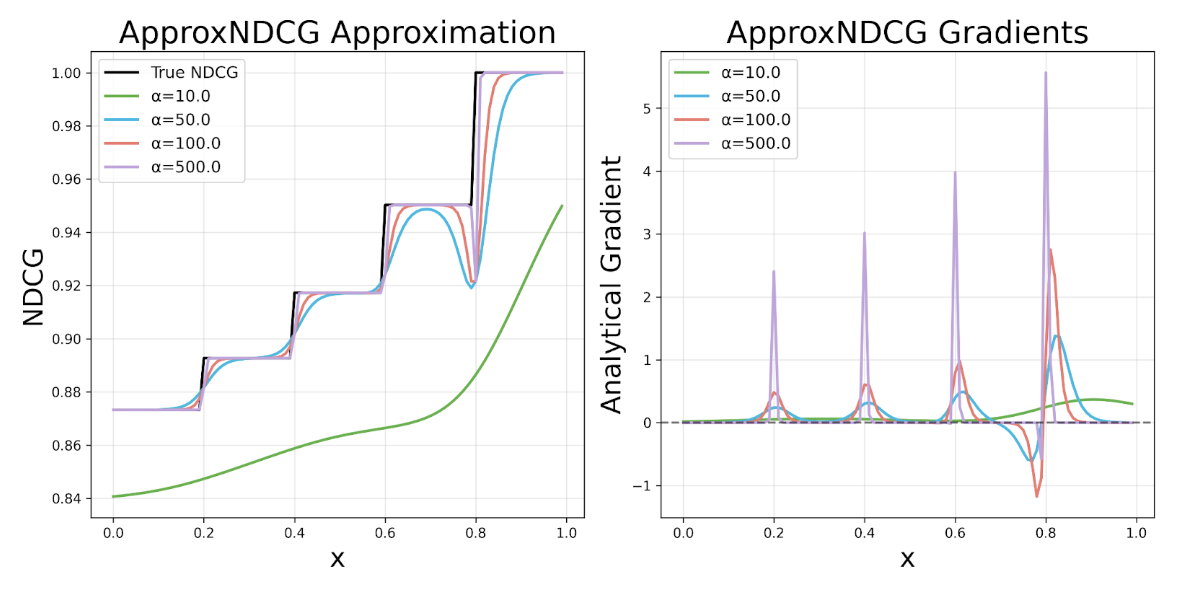}
    \caption{Given ground truth label $\psi=[1.0,0.8,0.6,0.4,0.2]$, the scores $\mathbf{s}=[x,0.8,0.6,0.4,0.2]$ and fix $\beta=0.1$, we visualize the ApproxNDCG Approximation Accuracy with different $\alpha$ and its corresponding gradients.}
    \label{fig_app_gradient}
\end{figure*}

We visualize the loss landscape of NeuralNDCG in \Cref{fig_ppa_gradient} and ApproxNDCG in \Cref{fig_app_gradient}. We find that with low $\tau$ and $\alpha$, the gradients become highly spiky and discontinuous, making optimization challenging and potentially unstable. In contrast, higher $\tau$ and $\alpha$ values yield smoother and more navigable gradients. However, when $\tau$ is too high (e.g., $\tau = 10.0$), the gradient curve becomes overly flat and the approximation error increases significantly, which can also hinder model performance. We observe this phenomenon consistently in both NDCG-based methods, NeuralNDCG and ApproxNDCG.

\subsection{Pseudo Code of Neural Sort}
\label{appendix_neural_sort_code}
The code for NeuralSort Eq \ref{neural_sort} is provided below.
\begin{verbatim}
import torch
import torch.nn.functional as F

def neuralsort(s, tau=1):
    #s.shape = [batch_size, list_size]
    s=s.unsqueeze(2)
    
    #A_s[i,j] = |s[i] - s[j]|
    A_s = s - s.transpose(1, 2)
    A_s = torch.abs(A_s)

    #B=A_s*ones
    n = s.size(1)
    one = torch.ones((n, 1),dtype=
        torch.float)
    B = torch.matmul(A_s, one @ 
        one.transpose(0, 1))
    
    #C=(n+1-2i)*s
    K = torch.arange(1, n + 1,dtype=
        torch.float)
    C = torch.matmul(s, (n + 1 - 2 * K)
        .unsqueeze(0))

    #P= softmax(((n+1-2i)*s-A_s*ones)/tau)
    P = (C - B).transpose(1, 2)
    P = F.softmax(P / tau, dim=-1)

    return P
\end{verbatim}
Given the score $\mathbf{s} = [9, 1, 5, 2]^T$ provided in Appendix \ref{appendix_sort}, we have

$$
\begin{aligned}
A_s=&
\begin{pmatrix}
0 & 8 & 4 & 7 \\
8 & 0 & 4 & 1 \\
4 & 4 & 0 & 3 \\
7 & 1 & 3 & 0 \\
\end{pmatrix},\\
B=&A_s\cdot \mathbf{1}_{k\times k}=
\begin{pmatrix}
19 & 19 & 19 &19 \\
13 & 13 & 13 & 13 \\
11 & 11 & 11 & 11 \\
11 & 11 & 11 & 11 \\
\end{pmatrix}\\
C=&[(n+1-2i)*\mathbf{s}]=
\begin{pmatrix}
27 & 9 & -9 & -27 \\
3 & 1 & -1 & -3 \\
15 & 5 & -5 & -15 \\
6 & 2 & -2 & -6 \\
\end{pmatrix}
\end{aligned}
$$
Based on Eq \ref{neural_sort}, we can get $\widehat{P}_{\text{sort}}(s)$
$$
\begin{aligned}
&\widehat{P}_{\text{sort}}(s)=\text{softmax }[\frac{C-B}{\tau}]=\\
&\begin{pmatrix}
0.98 & 1.5e-8 & 0.018 & 2.2e-6 \\
0.017 & 2.3e-3 & 0.93 & 0.047 \\
2.2e-7 & 0.26 & 0.035 & 0.71 \\
6.8e-14 & 0.73 & 3.3e-5 &0.27 \\
\end{pmatrix}
\end{aligned}
$$
Finally, we can get the permutation of the  sorted score vector as shown in \Cref{tab_p_hat}:
$$
\widehat{P}_{\text{sort}}(s)\cdot \mathbf{s}=
\begin{pmatrix}
8.9282 & 4.9420 & 1.8604 & 1.2691 \\
\end{pmatrix}
$$

We also show the sum of columns and rows:
$$
\text{column sum}:\begin{pmatrix}
0.9991 & 0.9928 & 0.9872 & 1.0208 \\
\end{pmatrix}
$$
$$
\text{row sum}:\begin{pmatrix}
1.0000 & 1.0000 & 1.0000 & 1.0000 \\
\end{pmatrix}^T
$$
As discussed above, $\widehat{P}_{\text{sort}}(s)$ is not column-stochastic, meaning each column may not sum to one. This can cause some $G(\psi_i)$ to contribute to the overall loss objective disproportionately and adversely affect model performance. The ablation studies are shown in \Cref{setup}.

\section{Training Details}
\label{appendix_train}
%Adhering to the settings in HuggingFace Alignment Handbook \citep{alignment_handbook2023}, we use a learning rate of $5 \times 10^{-7}$ and a total batch size of 128 for all training processes. The models are trained using the AdamW optimizer \citep{kingma2014adam} on 4 Nvidia V100-32G GPUs for Qwen2-0.5B models and 16 Nvidia V100-32G GPUs for Mistral-7B. 
The detailed training hyperparameters of Mistral-7B are shown in \Cref{tab_train_hyper}. 

\begin{table*}[h]
\small
\centering
\begin{tabular}{lccc}
\toprule
 \textbf{Datasets}& \textbf{Examples} & \textbf{Judge Model} & \textbf{Notes}  \\
\midrule
\textbf{UltraChat200k}&208k &-&SFT\\
\textbf{ListUltraFeedback}$_{\text{train}}$&59.9k &-&Permutative Preference Alignment\\
\midrule
\multirow{2}{*}{\textbf{ListUltraFeedback}$_{\text{test}}$} &\multirow{2}{*}{1968} & RLHFlow Pair-Preference & Pair-Preference win rates  \\
 & & ArmoRM & Scoring win rates  \\
\textbf{AlpacaEval} & 805 & GPT-4 Turbo & Pair-Preference win rates  \\
\textbf{MT-Bench} & 80 &  GPT-4 Turbo & Scoring win rates \\
\bottomrule
\end{tabular}
\caption{Details of training datasets and evaluation datasets.}
\label{tab_datasets}
\end{table*}

\begin{table}[h]
\small
\centering
\begin{tabular}{lc}
\toprule
 \textbf{Hyperparameters}& \textbf{value}  \\
\midrule
Mini Batch&1\\
Gradient Accumulation Steps&8\\
GPUs& 16$\times$Nvidia V100-32G\\
Total Batch Size&128\\
Learning Rate&5e-7\\
Epochs&1\\
Max Prompt Length&512\\
Max Total Length&1024\\
Optimizer&AdamW\\
LR Scheduler& Cosine\\
Warm up Ratio&0.1\\
Random Seed &42\\
\midrule
$\beta$&\{0.01, $\textbf{0.05}^*$, 0.1\}\\
$\tau$ for PPA & 1.0\\
$\alpha$ for ApproxNDCG &25\\
\midrule
Sampling Temperature&0\\
Pair-Preference Proxy Model&RLHFlow Pair-Preference\\
Scoring Proxy Model& ArmoRM\\
GPT Judge& GPT-4-Turbo\\
AlpacaEval Judge&alpaca\_eval\_gpt4\_turbo\_fn\\
\bottomrule
\end{tabular}
\caption{Training hyperparameters for Mistral-7B and Llama3.1-8B models.}
\label{tab_train_hyper}
\end{table}
Since Nvidia v100 is incompatible with the bf16 type, we use fp16 for mixed precision in deepspeed configuration. Notably, as the ListMLE method doesn't have normalization, it will encounter loss scaling errors with mixed precision settings.

\section{Supplementary Results}
\subsection{Proxy Models Results}
\label{appendix_sup_proxy}
The supplementary results of the Proxy Model Win Rate are shown in \Cref{tab_sup_beta} and \Cref{tab_sup_list}. For PPA, we fix $\tau=1.0$. For ApproxNDCG, we fix $\alpha=25$ because it is the parameter $\alpha\cdot\beta$ that controls the approximation accuracy of the sigmoid function in Eq \ref{approx_rank}.

\begin{table*}[h]
\small
    \centering
    \begin{tabular}{lcccccc}
        \toprule
        \textbf{Run Name} &$\beta$& \textbf{Pair-Preference} & \textbf{Scoring}&$\beta$& \textbf{Pair-Preference} & \textbf{Scoring} \\
        \midrule
        \multirow{2}{*}{Single Pair}&0.05& 57.24 & 54.04&0.01& 55.59 & 51.73\\
         &0.1& \textbf{60.75} & 56.86&0.5& 58.97 & \textbf{58.16}\\
        \cmidrule(lr){1-1}\cmidrule(lr){2-4} \cmidrule(lr){5-7}
        \multirow{2}{*}{BPR} &0.05& 59.86 & 56.86&0.01& 56.13 & 55.16 \\
         &0.1& \textbf{60.32} & \textbf{58.33}&0.5& 54.24 & 55.31 \\
        \cmidrule(lr){1-1}\cmidrule(lr){2-4} \cmidrule(lr){5-7}
        \multirow{2}{*}{All Pairs} &0.05& 62.12 & 58.36&0.01& 61.18 & 56.35\\
        &0.1& \textbf{63.82} & \textbf{60.54}&0.5& 56.12 & 55.77 \\
        \cmidrule(lr){1-1}\cmidrule(lr){2-4} \cmidrule(lr){5-7}
        \multirow{2}{*}{SLiC}&0.05& \textbf{63.31} & \textbf{60.70}&0.01& 59.30 & 55.61 \\
        &0.1& 62.68 & 60.34&0.5& 55.23 & 55.44 \\
        \cmidrule(lr){1-1}\cmidrule(lr){2-4} \cmidrule(lr){5-7}
        \multirow{2}{*}{LambdaRank}&0.05& 60.77 & 56.07&0.01& 54.52 & 51.35 \\
        &0.1& \textbf{62.30} & \textbf{59.04}&0.5& 57.72 & 56.71 \\
        \cmidrule(lr){1-1}\cmidrule(lr){2-4} \cmidrule(lr){5-7}

        \multirow{2}{*}{ListMLE}&0.05& 61.81 & 57.60&0.01& 57.49 & 55.16 \\
        &0.1& \textbf{63.03} & \textbf{59.76}&0.5& 56.05 & 55.77 \\
        \cmidrule(lr){1-1}\cmidrule(lr){2-4} \cmidrule(lr){5-7}

        \multirow{3}{*}{ApproxNDCG}&0.05& 58.66 & 54.34&0.01& 55.56 & 50.76\\
        &0.1& \textbf{61.46} & \textbf{58.59}&0.2& 60.04 & 57.27 \\
        &0.5& 58.71 & 57.39&1.0& 56.61 & 56.00 \\
        \cmidrule(lr){1-1}\cmidrule(lr){2-4} \cmidrule(lr){5-7}
        \multirow{2}{*}{PPA}&0.05& 63.92 & 60.09&0.01& 59.58 & 55.46 \\
        &0.1& \underline{\textbf{64.25}} & \underline{\textbf{61.36}}&0.5& 58.41 & 57.65 \\
        \bottomrule
    \end{tabular}
\caption{Supplementary Results across different $\beta$ on Qwen2-0.5B.}
    \label{tab_sup_beta}
\end{table*}

\begin{table*}[h]
\small
    \centering
    \begin{tabular}{lccccc}
        \toprule
        \textbf{Run Name} &\textbf{List Size}& \textbf{Pair-Preference} & \textbf{Scoring}& \textbf{Pair-Preference} & \textbf{Scoring} \\
        \midrule    
        &&\multicolumn{2}{c}{$\beta=0.1$}&\multicolumn{2}{c}{$\beta=0.05$}\\
        \midrule
        \multirow{4}{*}{All Pairs}&2&60.75&56.86&57.24&54.04\\
        &4&63.26&\textbf{60.90}&61.59&\textbf{58.54}\\
        &6&63.03&59.50&\textbf{62.83}&57.93\\
        &8&\textbf{63.82}&60.54&62.12&58.36\\
        \cmidrule(lr){1-2}\cmidrule(lr){3-4} \cmidrule(lr){5-6}
        \multirow{4}{*}{SLiC}&2&63.44&59.07&61.00&57.39\\
        &4&\textbf{63.79}&\underline{\textbf{61.40}}&\textbf{64.04}&60.54\\
        &6&63.64&61.15&62.01&58.61\\
        &8&62.68&60.34&63.31&\textbf{60.70}\\
        \cmidrule(lr){1-2}\cmidrule(lr){3-4} \cmidrule(lr){5-6}
        \multirow{4}{*}{LambdaRank}&2&60.85&57.62&59.76&56.02\\
        &4&61.10&58.05&59.88&55.51\\
        &6&62.09&57.72&\textbf{62.02}&\textbf{56.81}\\
        &8&\textbf{62.30}&\textbf{59.04}&60.77&56.07\\
        \cmidrule(lr){1-2}\cmidrule(lr){3-4} \cmidrule(lr){5-6}
        \multirow{4}{*}{ListMLE}&2&60.14&57.01&57.14&53.53\\
        &4&\textbf{63.57}&\textbf{61.23}&\textbf{61.94}&\textbf{58.49}\\
        &6&62.78&60.92&61.18&57.83\\
        &8&63.03&59.76&61.81&57.60\\
        \cmidrule(lr){1-2}\cmidrule(lr){3-4} \cmidrule(lr){5-6}
        \multirow{4}{*}{ApproxNDCG}&2&59.73&57.72&\textbf{61.56}&\textbf{58.26}\\
        &4&59.65&56.45&60.11&55.79\\
        &6&60.70&57.32&59.53&56.35\\
        &8&\textbf{61.46}&\textbf{58.59}&58.66&54.34\\
        \cmidrule(lr){1-2}\cmidrule(lr){3-4} \cmidrule(lr){5-6}
        \multirow{4}{*}{PPA}&2&61.94&58.00&58.69&55.89\\
        &4&62.91&59.96&62.65&58.56\\
        &6&64.02&60.11&61.08&59.43\\
        &8&\underline{\textbf{64.25}}&\textbf{61.36}&\textbf{63.92}&\textbf{60.09}\\
        \bottomrule
    \end{tabular}
    \caption{Supplementary Results across different list sizes on Qwen2-0.5B. In practice, we keep the response with the highest label and the one with the lowest label, then conduct random sampling from the remaining responses.}
    \label{tab_sup_list}
\end{table*}

\subsection{Supplementary Results for Mistral-7B}
\label{appendix_sup_mistral}
\begin{table*}[h]
\small
    \centering
    \begin{tabular}{lcccccc}
        \toprule
        \multirow{3}{*}{\textbf{\vspace{6pt}Method}}  & \multirow{3}{*}{\textbf{\vspace{6pt}Type}}& \multicolumn{2}{c}{\textbf{Proxy Model}} & \multicolumn{2}{c}{\textbf{General Benchmark}} & \multirow{3}{*}{\textbf{\vspace{6pt}Avg.}}\\
        \cmidrule(lr){3-4} \cmidrule(lr){5-6}
        &&  \textbf{Pair-Preference} &\textbf{Scoring} &   \textbf{AlpacaEval} & \textbf{MT-Bench} &\\
        \midrule
        Single Pair&Pairwise&71.90&70.66& 74.75& 52.19&\cellcolor{gray!20}67.38\\
        BPR&Pairwise&84.43&82.37& 86.69& 63.44&\cellcolor{gray!20}79.23\\
        Others vs Worst&Pairwise&82.95&80.84& 84.64&62.78&\cellcolor{gray!20}77.80\\
        All Pairs &Pairwise&\underline{\textbf{85.34}}&83.31& 82.79& 61.56&\cellcolor{gray!20}78.25\\
        SLiC&Pairwise&84.12&83.46& 83.27& 66.25&\cellcolor{gray!20}79.28\\
        LambdaRank&Listwise&85.11&82.52&86.13&\underline{\textbf{69.06}}&\cellcolor{gray!20}80.71\\
        ListMLE&Listwise&83.79&\underline{\textbf{83.61}}& 83.46& 66.56&\cellcolor{gray!20}79.35\\
        \midrule
        ApproxNDCG&Listwise&82.04&74.64&85.80& 67.50&\cellcolor{gray!20}77.50\\
        PPA&Listwise&84.98&83.05& \underline{\textbf{87.54}}& 67.81&\cellcolor{gray!20}\underline{\textbf{80.85}}\\
        \bottomrule
    \end{tabular}
\caption{PPA outperforms other baselines on win rates of aligned Mistral-7B against Zephyr-7B-SFT. We set $\beta=0.01$ for Single Pair and $\beta=0.05$ for other approaches to achieve the best performance. The other settings are the same as in \Cref{tab: main results}.}
\label{mistral7b}
\end{table*}

We observe that decreasing the hyperparameter $\beta$ may increase the performance when language models scale up to 7B parameters. All methods achieve their best performance with $\beta=0.05$ except for Single Pair with $\beta=0.01$. Our approach PPA consistently achieves the best overall performance, shown in \Cref{tab_sup_mistral}.

\begin{table*}[h]
\small

\centering
\begin{tabular}{lcccc}
    \toprule
    \textbf{Method} & $\beta$&\textbf{Pair-Preference}& \textbf{Scoring}& \textbf{AlpacaEval}  \\
    \midrule
    Single Pair &\multirow{8}{*}{0.1}&61.26&70.21&64.45\\
    BPR &&79.73&77.59& 78.39\\
    All Pairs    & & 79.22   & 78.43& 77.65  \\
    SLiC &&76.17&75.36& 73.04\\
    LambdaRank &&80.82&78.53& 81.01\\
    ListMLE &&78.58&79.22& 75.12\\
    ApproxNDCG& &76.12&69.21& \textbf{82.50}\\
    PPA   & &  \textbf{83.13}&\textbf{81.66}& 81.07\\
    \midrule
    Single Pair&\multirow{10}{*}{0.05} &66.44&65.50& 68.87\\
    BPR&&84.43&82.37& 86.69\\
    Others vs Worst&&82.95&80.84& 84.64\\
    All Pairs& &85.34&83.31& 82.79\\
    %BCE&&53.76&54.17& -\\
    RankNet&&85.61&84.22& 86.94\\
    %ListNet&&54.80&56.21& -\\
    SLiC&&84.12&83.46& 83.27\\
    LambdaRank&&85.11&82.52&86.13\\
    ListMLE&&83.79&83.61& 83.46\\
    ApproxNDCG&&82.04&74.64&85.80\\
    PPA&&  84.98&83.05& \underline{\textbf{87.54}}\\
    \midrule
    Single Pair&\multirow{4}{*}{0.01} &71.90&70.66& 74.75\\
    BPR&&77.01&78.46& 86.71\\
    All Pairs&&72.66&74.09& 82.44\\
    PPA& & 73.17&75.00&84.51\\
    \bottomrule
    \end{tabular}
\caption{Model Scale Up results in Mistral-7B.}
\label{tab_sup_mistral}
\end{table*}

\begin{table*}[h]
\small
    \centering
    \setlength{\tabcolsep}{4pt}
    \begin{tabular}{lcccc}
        \toprule
        \textbf{Method} & \textbf{Pair-Preference }& \textbf{Scoring }& \textbf{AplacaEval} & \textbf{Arena-Hard}\\
        \midrule
        All Pairs     & 72.96   & 74.39&59.64& \underline{\textbf{61.64}}\\
        SLiC&72.84&75.04&60.20&59.17\\
        ListMLE &72.46&74.77&59.83&55.18\\
        LiPO&71.80&72.74&60.38&59.69\\
        \midrule
        PPA    & \underline{\textbf{74.34}}  & \underline{\textbf{75.58}}&\underline{\textbf{61.32	}}&60.17\\
        \bottomrule
    \end{tabular}
\caption{Model Scale Up: PPA outperforms other approaches on Llama3.1-8B.
}
\label{app_tab:llama3-8b}
\end{table*}

\begin{figure*}[h]
    \centering
    \includegraphics[width=\textwidth]{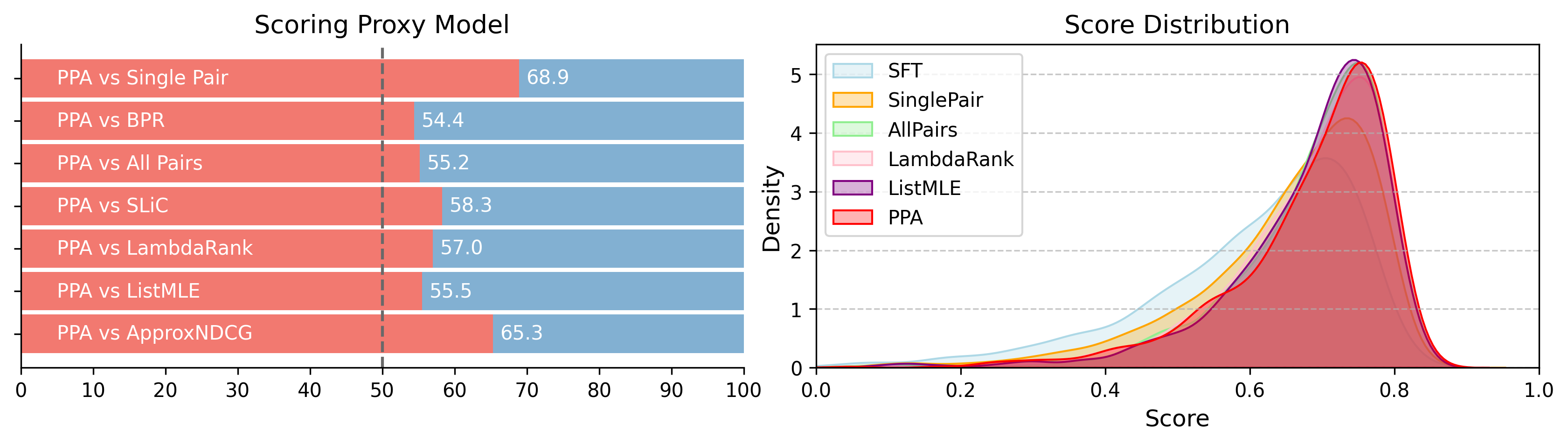}
    \caption{PPA demonstrates superior performance compared to other methods in Scoring Proxy model win rates on Mistral-7B, while also shifting the distribution of response reward scores more significantly to the right (i.e., increasing reward scores).}
    \label{fig_sup_compare}
\end{figure*}

To further explore the distribution shift during human preference alignment, we demonstrate the score distribution of all methods of which scores are assigned by the Reward model ArmoRM \citep{ArmoRM} in \Cref{fig_sup_compare}. The PPA method causes the reward score distribution to shift more significantly to the right, resulting in fewer instances at lower scores. Consequently, when compared to the SFT model, its win rate is not as high as methods like All Pairs, SLiC, and ListMLE. However, it can outperform these methods in direct comparisons.

\section{ApproxNDCG Analysis}
\label{appendix_approx}
In ApproxNDCG, we observe similar results to NeuralNDCG; the model achieves optimal performance only when the approximation accuracy reaches a certain threshold. First, we prove that the Accuracy of ApproxNDCG is relevant to the multiplication of $\alpha$ and $\beta$ when we employ the score function in Eq \ref{scores_eq}.

First, let's define the term for the probability ratio as $R_{ji}$:
\[
R_{ji} = \frac{\pi_\theta(y_j|x)\pi_{ref}(y_i|x)}{\pi_{ref}(y_j|x)\pi_\theta(y_i|x)}
\]
Note that from the properties of logarithms, we have:
\[
\log(R_{ji}) = \log\frac{\pi_\theta(y_j|x)}{\pi_{ref}(y_j|x)} - \log\frac{\pi_\theta(y_i|x)}{\pi_{ref}(y_i|x)}
\]
Using this simplification, the derivation becomes much more compact:

\begin{equation}
\small
\begin{aligned}
\widehat{\tau(j)} &= 1+\sum_{i\ne j}\frac{\exp{(-\alpha(s_j-s_i))}}{1+\exp{(-\alpha(s_j-s_i))}} \\
&= 1+\sum_{i\ne j}\frac{1}{1+\exp(\alpha(s_j-s_i))} \\
&= 1+\sum_{i\ne j}\frac{1}{1+\exp(\alpha\beta\log R_{ji})} \\
&= 1+\sum_{i\ne j}\frac{1}{1+ (R_{ji})^{\alpha\beta}} \\
&= \dots
\end{aligned}
\label{approx_rank2}
\end{equation}

\begin{figure*}[h]
    \centering
    \includegraphics[width=\textwidth]{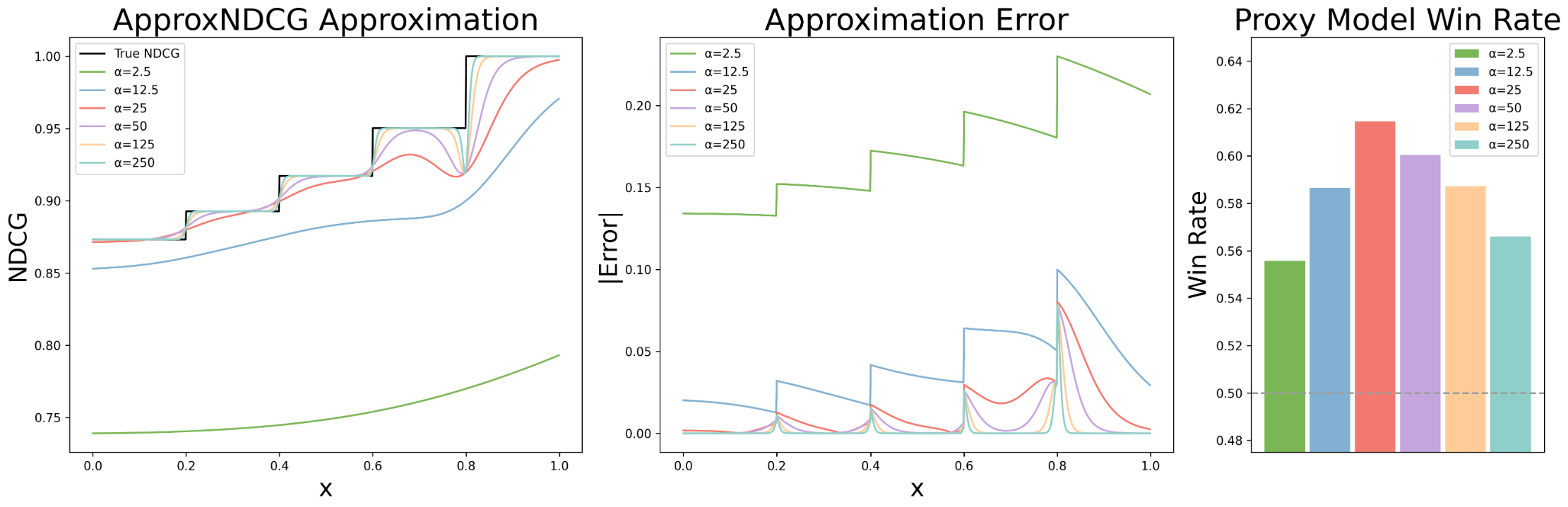}
    \caption{Given ground truth label $\psi=[1.0,0.8,0.6,0.4,0.2]$, the scores $\mathbf{s}=[x,0.8,0.6,0.4,0.2]$ and fix $\beta=0.1$, we visualize the ApproxNDCG Approximation Accuracy with different $\alpha$ and its corresponding absolute value of error and Pair-Preference proxy model win rate against SFT model.}
    \label{fig_approx}
\end{figure*}

Then, we illustrate the Approximation accuracy and model performance of ApproxNDCG with different hyperparameters $\alpha\cdot\beta$ in \Cref{fig_approx}.

Notice that the approximation accuracy of ApproxNDCG decreases as $\alpha$ increases, which is opposite to NeuralNDCG.

\section{Training Efficiency}
The computational complexity of each method depends on evaluating $\pi_{\theta}(y_j|x)$ and $\pi_{\text{ref}}(y_j|x)$ for each $y_j\in \{\mathbf{Y}\}$ to get corresponding scores in Eq \ref{scores_eq}, which is $\mathcal{O}(K)$, where K is the list size of multiple responses. Subsequently, the pairwise comparison of multiple responses can be efficiently computed using PyTorch's broadcasting mechanism to perform matrix subtraction.The resulting matrix $P[i,j]$ represents the value of $s_i-s_j$. Therefore, for pairwise methods, it suffices to consider only the upper triangular matrix, excluding diagonal elements. This approach does not significantly increase training time when performing pairwise comparisons. The training times and GPU memory usage of Mistral-7B and Qwen2-0.5B models are shown in \Cref{tab_train_time} and \Cref{tab_train_time_0.5}.
 
\begin{table*}[h]
\small
\centering
\begin{tabular}{lccc}
\toprule
 \textbf{Run Name}& \textbf{List Size}&\textbf{Training Time}&\textbf{GPU Memory Usage}  \\
\midrule
Single Pair&2&3h 28m&92.44\%\\
BPR&8&12h 42m&93.43\%\\
All Pairs&8&12h 38m&93.63\%\\
SLiC&8&11h 42m&93.79\%\\
LambdaRank&8&12h 14m&93.66\%\\
ListMLE&8&12h 26m&93.29\%\\
\midrule
ApproxNDCG&8&12h 56m&93.64\%\\
PPA&8&11h 39m&93.73\%\\
\bottomrule
\end{tabular}
\caption{Training Time and GPU memory usage on 16$\times$Nvidia V100-32G with Mistral-7B.}
\label{tab_train_time}
\end{table*}

\begin{table*}[h]
\small
\centering
\begin{tabular}{lccccc}
\toprule
\textbf{List Size}& \textbf{Approach}&\textbf{Training Time}& \textbf{List Size}& \textbf{Approach}&\textbf{Training Time} \\
\cmidrule(lr){1-3}\cmidrule(lr){4-6}
\multirow{5}{*}{2}&PPA&3h24m&\multirow{5}{*}{4}&PPA&4h27m\\
&DPO&3h17m&&DPO&4h10m\\
&ListMLE&3h38m&&ListMLE&4h23m\\
&LiPO&3h32m&&LiPO&4h47m\\
&SLiC&3h28m&&SLiC&4h14m\\
\cmidrule(lr){1-3}\cmidrule(lr){4-6}
\multirow{5}{*}{6}&PPA&6h31m&\multirow{5}{*}{8}&PPA&7h23m\\
&DPO&6h12m&&DPO&7h14m\\
&ListMLE&6h56m&&ListMLE&7h34m\\
&LiPO&6h17m&&LiPO&7h57m\\
&SLiC&6h24m&&SLiC&8h04m\\
\bottomrule
\end{tabular}
\caption{Training Time and GPU memory usage on Qwen2-0.5B.}
\label{tab_train_time_0.5}
\end{table*}

NeuralNDCG (as used in PPA) requires NeuralSort and Sinkhorn scaling for each list. In our experiments \Cref{tab_sort_time}, we use lists of size $n=8$, thus operating on an $8 \times 8$ matrix for each sample. The computational complexity per sample is $O(n^2 \cdot \text{iter})$, where $\text{iter}$ is the number of Sinkhorn scaling iterations (set to 50 in our experiments). In contrast, DPO operates on all pairs within a list, with complexity $O(n^2)$ per sample.

\begin{table*}[h]
\small
\centering
\begin{tabular}{lccc}
\toprule
 \textbf{Method}& \textbf{Complexity}&\textbf{Time}&\textbf{Time per sample}  \\
\midrule
DPO&$O(n^2)$&7.394s&0.739ms\\
PPA&$O(50\cdot n^2)$&368.5s&36.850ms\\
\bottomrule
\end{tabular}
\caption{Wall-clock time of sorting operations for 10,000 samples on DPO and PPA.}
\label{tab_sort_time}
\end{table*}

It is important to note that specific training times and GPU memory usage can exhibit random fluctuations. This result is intended to demonstrate that the training times for pairwise and listwise methods on multiple responses with the same list size do not show significant differences.

\section{Response Samples}
\label{appendix_sample}
The average response length on the AlpacaEval dataset is shown in \Cref{tab_length}. We also provide response samples in \Cref{tab_samples} from each baseline and our method PPA demonstrates more details in generated responses.

\begin{table*}[h]
\caption{Average response length on AlpacaEval dataset.}
\label{tab_length}
\centering
\resizebox{\textwidth}{!}{%
\begin{tabular}{lccccccccc}
\toprule
\textbf{Method}&SFT&Single Pair&BPR&All Pairs&SLiC&LambdaRank&ListMLE&ApproxNDCG&NeuralNDCG\\
\midrule
\textbf{Length}&828&877&1096&1054&1023&1140&1041&1284&1120\\
\bottomrule
\end{tabular}
}
\end{table*}

\begin{table*}[h]
\centering
\caption{Response samples from AlpacaEval where PPA outperforms other methods in quality judged by GPT-4 Turbo.}
\label{tab_samples}
\resizebox{0.8\textwidth}{!}{%
\begin{tabular}{lp{10cm}}
\toprule
\textbf{\texttt{Prompt}} & \texttt{Q: Who is Larry Page?} \\
\midrule
\textbf{\texttt{PPA}} & \texttt{Larry Page is a co-founder, former CEO, and current CEO of Alphabet Inc., the parent company of Google. He was born on March 26, 1973, in East Lansing, Michigan, and graduated from Stanford University with a degree in computer science and artificial intelligence. Page, along with Sergey Brin, developed the PageRank algorithm, which forms the basis of Google's search engine. In 1998, they founded Google, which quickly became one of the most successful and influential technology companies in the world. Page served as Google's CEO from 1999 to 2001 and again from 2011 to 2015, during which time he oversaw the company's expansion into new markets, including mobile devices, cloud computing, and artificial intelligence. In 2015, Page stepped down as CEO of Google and became the CEO of Alphabet, the holding company that was created to separate Google's core internet business from its other ventures. As Alphabet's CEO, Page has focused on developing new technologies and expanding into emerging markets, such as healthcare and autonomous vehicles. Page is widely regarded as one of the most innovative and influential entrepreneurs of our time, and his contributions to the field of computer science and technology have had a profound impact on the world.} \\
\bottomrule
\end{tabular}%
}
\end{table*}

\begin{table*}[h]
\centering
\resizebox{0.8\textwidth}{!}{%
\begin{tabular}{lp{10cm}}
\toprule
\textbf{\texttt{SFT}} & \texttt{Larry Page is the co-founder and former CEO of Google, a multinational technology company. He is also the co-founder of Alphabet Inc., the parent company of Google. Page is known for his contributions to the development of the PageRank algorithm, which is used to determine the importance of websites in search engine rankings. He is also involved in various philanthropic efforts, including the creation of the Google Foundation, which supports projects related to education, health, and the environment.} \\
\midrule
\textbf{\texttt{Single Pair}} & \texttt{Larry Page is an American entrepreneur, computer scientist, and co-founder of Google, along with Sergey Brin. He was born on March 26, 1973, in East Lansing, Michigan, and graduated from Stanford University with a degree in computer science and artificial intelligence. Page is known for his innovative ideas and leadership in the technology industry, particularly in the areas of search engines, artificial intelligence, and renewable energy. He currently serves as the CEO of Alphabet Inc., the parent company of Google, and is considered one of the most influential figures in the tech world.} \\
\midrule
\textbf{\texttt{BPR}} & \texttt{Larry Page is a co-founder of Google, along with Sergey Brin. He was born on March 26, 1973, in East Lansing, Michigan, USA. Page and Brin started Google in 1998 while they were both PhD students at Stanford University. Page's primary role in the company has been as CEO from 2001 to 2019, and he currently serves as the CEO of Alphabet Inc., the parent company of Google. Under his leadership, Google has become one of the most successful and influential technology companies in the world, with a market capitalization of over \$1 trillion. Page is also known for his philanthropic efforts, including funding research on renewable energy and supporting education initiatives.} \\
\midrule
\textbf{\texttt{All Pairs}} & \texttt{Larry Page is a co-founder of Google, along with Sergey Brin. He was born on March 26, 1973, in East Lansing, Michigan, USA. Page and Brin started Google in 1998 while they were both PhD students at Stanford University. Page is known for his focus on innovation, simplicity, and user experience, and he has been instrumental in shaping Google's vision and strategy. He served as CEO of Alphabet Inc., the parent company of Google, from 2015 to 2019, and currently serves as the CEO of Alphabet's subsidiary, Google. Page is also a philanthropist and has been involved in various charitable initiatives, including the PageRipples Foundation, which supports education and environmental causes.} \\
\bottomrule
\end{tabular}%
}
\end{table*}

\begin{table*}[h]
\small
\centering
\resizebox{0.8\textwidth}{!}{%
\begin{tabular}{lp{10cm}}
\toprule
\textbf{\texttt{SLiC}} & \texttt{Larry Page is a co-founder of Google, along with Sergey Brin. He was born on March 26, 1973, in East Lansing, Michigan, USA. Page and Brin started Google in 1998 while they were both PhD students at Stanford University. Page is currently the CEO of Alphabet Inc., the parent company of Google, and is responsible for overseeing the company's overall strategy and direction. He is also a prominent figure in the technology industry, known for his focus on innovation, sustainability, and philanthropy.} \\
\midrule
\textbf{\texttt{LambdaRank}} & \texttt{Larry Page is a co-founder of Google, along with Sergey Brin. He was born on March 26, 1973, in East Lansing, Michigan, USA. Page and Brin started Google in 1998 while they were both PhD students at Stanford University. Page served as the CEO of Google from 1999 to 2001, and then again from 2011 to 2015. He currently serves as the CEO of Alphabet Inc., the parent company of Google. Page is known for his focus on innovation, sustainability, and philanthropy, and has been recognized as one of the most influential people in the world.} \\
\midrule
\textbf{\texttt{ListMLE}} & \texttt{Larry Page is a co-founder of Google, along with Sergey Brin. He was born on March 26, 1973, in East Lansing, Michigan, USA. Page and Brin started Google in 1998 while they were both PhD students at Stanford University. Page is currently the CEO of Alphabet Inc., the parent company of Google, and is responsible for overseeing the company's overall strategy and direction. He is also a prominent figure in the technology industry, known for his focus on innovation, sustainability, and philanthropy.} \\
\midrule
\textbf{\texttt{ApproxNDCG}} & \texttt{Larry Page is a co-founder of Google, along with Sergey Brin. He was born on March 26, 1973, in East Lansing, Michigan, and grew up in the Detroit area. Page and Brin met at Stanford University, where they developed the PageRank algorithm, which became the basis for Google's search engine.\textbackslash n\textbackslash n Page served as the CEO of Google from 1999 to 2001, and then again from 2011 to 2019. During his tenure, he oversaw the company's growth and expansion, including the development of new products and services, such as Google Maps, Google Earth, and Google Glass.\textbackslash n\textbackslash n Page is known for his focus on innovation, user experience, and sustainability. He has also been involved in various philanthropic efforts, including the creation of the Google.org foundation and the development of the Google Self-Driving Car Project.\textbackslash n\textbackslash n Today, Page is the CEO of Alphabet, the parent company of Google, and continues to play a key role in shaping the future of technology and the internet.} \\
\bottomrule
\end{tabular}%
}
\end{table*}

\end{document}